\documentclass{article}

\PassOptionsToPackage{numbers,sort&compress}{natbib}
\usepackage[eandd,preprint]{neurips_2026}

\usepackage[utf8]{inputenc}
\usepackage[T1]{fontenc}
\usepackage{hyperref}
\usepackage{url}
\usepackage{booktabs}
\usepackage{amsfonts}
\usepackage{amsmath,amssymb}
\DeclareMathOperator{\softmax}{softmax}

\usepackage{nicefrac}
\usepackage{microtype}
\usepackage[table]{xcolor}
\usepackage{graphicx}
\usepackage{placeins}
\usepackage{caption}
\usepackage{enumitem}
\setlist{itemsep=1pt,parsep=0pt,topsep=2pt,partopsep=0pt}
\setlist[enumerate]{itemsep=2pt,parsep=0pt,topsep=2pt,partopsep=0pt}
\usepackage{tikz}
\usetikzlibrary{arrows.meta,positioning,fit,backgrounds,shapes.geometric,decorations.pathreplacing}
\usepackage{pgfplots}
\pgfplotsset{compat=1.18}

\newlength{\energytilew}

\title{When Accuracy Is Not Enough: Uncertainty Collapse between Noisy Label Learning and Out-of-Distribution Detection}

\author{%
  Ningkang Peng \\
  Nanjing Normal University
  \And
  Jingyang Mao \\
  Nanjing Normal University
  \And
  Runhan Zhou \\
  Nanjing Normal University
  \And
  Peirong Ma \\
  Nanjing Normal University
  \And
  Yanhui Gu \\
  Nanjing Normal University
}

\begin{document}
\flushbottom

\maketitle

\begin{abstract}
Learning with noisy labels (LNL) is typically benchmarked by closed-set classification accuracy, yet deployment often requires classifiers to reject out-of-distribution (OOD) inputs.
We present a learner-agnostic ACC--OOD benchmark that freezes LNL checkpoints and evaluates them with standardized near-/far-OOD routing and post-hoc scores across synthetic and real label noise.
The benchmark reveals a recurring failure mode: high closed-set accuracy does not ensure OOD reliability, because low-confidence, misclassified in-distribution samples can overlap the score and feature regions occupied by OOD inputs under noisy training.
We term this pathology \emph{uncertainty collapse}.
This structural overlap can make high-accuracy LNL methods lose separability at the ID-error/OOD interface under standard OOD scores.
As an intervention, we study Virtual Margin Regularization (VMR), a lightweight repair probe demonstrated mainly with PSSCL that synthesizes boundary virtual outliers on trusted ID batches and widens the energy margin.
VMR partially reduces the collapse-induced far-OOD failure without replacing the host objective or sacrificing closed-set accuracy in the tested settings.
These results support LNL benchmarks that co-report closed-set generalization, open-world reliability, and structural overlap diagnostics.
\end{abstract}

\providecommand{\Cscore}{\mathcal{C}_{\mathrm{score}}}

\section{Introduction}
\label{sec:intro}

Learning with noisy labels (LNL) aims to recover useful decision boundaries from imperfect supervision, and has developed into a broad literature spanning robust losses, sample selection, co-training, contrastive filtering, and semi-supervised refinement~\cite{song2022lnlsurvey,zhang2018gce,han2018coteaching,dividemix,unicon}.
Most LNL benchmarks measure this goal through closed-set accuracy on clean in-distribution (ID) test data.
This evaluation is necessary, but it omits what the same trained classifier must do after deployment: once frozen, it may receive valid ID inputs, ambiguous or mislabeled ID cases, and inputs from unknown classes.
In that setting, the model must support rejection, abstention, or triage rather than always forcing a closed-set prediction~\cite{bendale2016openmax,geifman2017selective}.
The question for LNL is therefore not only whether noisy supervision can recover ID accuracy, but whether the recovered representation still separates ID uncertainty from genuine novelty.

Post-hoc OOD detection is the standard way to evaluate this open-world behavior.
Given a frozen classifier, scores such as maximum softmax probability, ODIN, Mahalanobis distance, and Energy use its confidence, logits, or features to reject inputs outside the training label space~\cite{hendrycks2017baseline,liang2018odin,lee2018mahalanobis,liu2020energy}.
These scores are often evaluated after clean supervised training, but LNL algorithms can reshape the same confidence and feature geometry through sample selection, relabeling, co-training, contrastive filtering, and semi-supervised refinement.
As a result, a method can improve closed-set accuracy while changing the score regions on which OOD rejection depends.
Recent work has begun to expose this interface: label noise can degrade post-hoc OOD detectors, and noisy data can disrupt the usual alignment between ID accuracy and OOD generalization~\cite{noisyelephant,wrongline}.
These observations point to an unresolved benchmark-level issue: OOD degradation may be a recurring consequence of representation-geometry changes induced by modern LNL training.
We study this behavior across LNL paradigms, noise sources, and near-/far-OOD routes, then use the resulting failures to diagnose the representation-level mechanism behind the ACC--OOD decoupling.

We present a \emph{learner-agnostic} ACC--OOD benchmark with standardized dataset routing and post-hoc scoring logic.
We freeze each LNL checkpoint and co-report closed-set accuracy with near-/far-OOD diagnostics across synthetic noise, human re-annotations (CIFAR-N), and large-scale real-world corruption (Food-101N and Animal-10N).
The benchmark reveals ACC--OOD reversals: LNL learners can retain competitive ID accuracy while standard OOD scores lose separability.
We trace this failure to a structural pathology we term \emph{uncertainty collapse}.
Operationally, collapse occurs when low-confidence ID-wrong samples become difficult to separate from OOD inputs under a fixed post-hoc score, as measured by ID-wrong/OOD AUROC and paired score-distribution diagnostics.
Under label noise, we observe a distinct low-confidence, misclassified ID subpopulation whose score and feature distributions overlap the support of OOD inputs.
OOD detection is then confounded not by a simple calibration error, but by the inability to separate ID errors from genuine novelty.

This diagnosis motivates a targeted plug-in intervention.
We study Virtual Margin Regularization (VMR) as a host-compatible repair probe, demonstrated mainly on PSSCL checkpoints, that regularizes the collapsed ID-error/OOD interface.
Rather than training a separate OOD detector or relabeling misclassified ID samples, VMR opens a geometric margin before post-hoc scoring.
Its far-OOD gains at matched closed-set accuracy suggest that uncertainty collapse is not only a benchmark observation, but also a partially recoverable failure mode that future LNL methods should be evaluated against.

Our main contributions are threefold:
\begin{itemize}[leftmargin=1.1em]
\item \textbf{A unified ACC--OOD benchmark suite.} Under standardized dataset routing and scoring logic, we co-report closed-set accuracy and open-world reliability in a learner-agnostic protocol spanning synthetic noise, human re-annotations (CIFAR-N), and large-scale real-world corruption (Food-101N, Animal-10N), revealing that high-accuracy LNL learners can fail at OOD detection under stress.
\item \textbf{Mechanistic diagnosis of uncertainty collapse.} Using a five-group taxonomy and clean-label comparisons inside the same pipeline, we connect ACC--OOD decoupling to a structural pathology where low-confidence, misclassified ID samples occupy score and feature regions shared with OOD inputs.
\item \textbf{A VMR plug-in for probing repair.} We study Virtual Margin Regularization (VMR), a host-compatible repair probe that synthesizes boundary virtual outliers from trusted ID batches and widens the far-OOD energy margin without replacing the LNL objective or retraining a detector.
\end{itemize}

\noindent\textbf{Repository.} Code and benchmark materials are available at \url{https://anonymous.4open.science/r/Anonymous-124F/}.

\section{Unified ACC--OOD Benchmark Protocol}
\label{sec:setup}

\paragraph{Protocol and positioning.}
Modern noisy-label learning (LNL) spans robust losses, sample selection, semi-supervised refinement, contrastive filtering, and curriculum mixing, yet standard benchmarks mostly report closed-set accuracy on corrupted ID test data.
We instead evaluate each LNL learner under a unified ACC--OOD protocol that treats the trained model as both a classifier and a frozen representation-and-score generator for open-world reliability.
This differs from designing a new OOD detector or an open-set noisy-label learner: prior studies show that label noise can harm OOD detection and OOD generalization~\cite{noisyelephant,wrongline}, while open-set LNL methods explicitly model out-of-class training samples~\cite{opensetlnl}; here we audit standard LNL checkpoints and diagnose the ID-error/OOD interface.
Given a learner $f_\theta$, we train it under its standard closed-set LNL objective, freeze the checkpoint, and compute all OOD scores post hoc, with no OOD data used for training, detector fitting, threshold tuning, or checkpoint selection.

\paragraph{Dataset and learner coverage.}
The benchmark covers synthetic noise, human annotation noise, and real-world noisy datasets.
For synthetic stress tests, we evaluate CIFAR-10 and CIFAR-100~\cite{krizhevsky2009learning}: CIFAR-10 uses symmetric noise rates $0.2$, $0.5$, $0.8$, and $0.9$ and asymmetric noise $0.4$, while CIFAR-100 uses symmetric rates $0.2$ and $0.5$.
For human and real-noise validation, we evaluate CIFAR-N variants~\cite{cifarn}, Food-101N~\cite{lee2018cleannet}, and Animal-10N~\cite{song2019selfie}; OOD evaluation is split into near-OOD cross-CIFAR neighbors and a fixed far-OOD suite of SVHN, LSUN-resize, Places365, DTD Textures, and MNIST.
Seven representative learners cover sample-selection/co-training, contrastive, and regularization or hybrid families; method-level defaults and hyperparameters are omitted in this arXiv preprint version.

\paragraph{Scores, diagnostics, and reporting.}
For each frozen checkpoint, we compute Energy~\cite{liu2020energy} as the primary post-hoc score, with MSP, MaxLogit, margin, entropy, and detector-control probes such as ODIN, Mahalanobis, ReAct, DNN, and ViM-style variants as robustness checks~\cite{hendrycks2017baseline,liang2018odin,lee2018mahalanobis,sun2021react,sun2022dnn,wang2022vim}.
All scores are converted to a common OOD-larger direction before AUROC and FPR95 computation, and the benchmark co-reports closed-set accuracy with near- and far-OOD reliability.
Calibration metrics such as ECE and NLL are supporting diagnostics, not substitutes for OOD reliability.
All structural diagnostics used later in Section~\ref{sec:pathology} are computed only after evaluation and are never used for training, threshold tuning, checkpoint selection, or model ranking.
Score-family and detector-control checks for score dependence, together with extended FPR95 material for the clean-vs-noise checkpoint study, are omitted in this arXiv preprint version.

\begin{table*}[!htbp]
\centering
\caption{\textbf{Unified ACC+OOD benchmark.}
Columns are dataset/noise settings; each method occupies five co-primary metric rows: ACC$\uparrow$, Near-OOD AUROC$\uparrow$, Near-OOD FPR95$\downarrow$, Far-OOD AUROC$\uparrow$, and Far-OOD FPR95$\downarrow$.
Bold marks the best value per column and underline marks the runner-up.
Results are single-run measurements intended to expose cross-method ACC--OOD patterns rather than definitive leaderboard rankings.
Unusually low-ACC cells are retained for transparency and treated as implementation/protocol stress cases rather than standalone method judgments.}
\label{tab:main-benchmark}
\scriptsize
\setlength{\tabcolsep}{3.0pt}
\renewcommand{\arraystretch}{0.92}
\resizebox{\textwidth}{!}{%
\begin{tabular}{@{}ll *{7}{c}@{}}
\toprule
\multicolumn{2}{c}{Dataset} & \multicolumn{5}{c}{CIFAR-10} & \multicolumn{2}{c}{CIFAR-100} \\
\cmidrule(lr){3-7}\cmidrule(lr){8-9}
Method & Metric
& Sym 0.2 & Sym 0.5 & Sym 0.8 & Sym 0.9 & Asym 0.4 & Sym 0.2 & Sym 0.5 \\
\midrule
\textbf{DivideMix} & Accuracy$\uparrow$ & 96.1 & 94.6 & 93.2 & 76.0 & 93.4 & 77.3 & 75.6 \\
& Near AUROC$\uparrow$ & 81.6 & 82.4 & 81.2 & 54.8 & 76.4 & \textbf{77.9} & \underline{76.8} \\
& Near FPR95$\downarrow$ & 62.1 & 51.1 & 64.7 & 93.7 & 72.7 & 80.6 & 80.7 \\
& Far AUROC$\uparrow$ & 69.7 & 77.2 & 77.9 & 49.2 & 49.2 & 63.1 & 60.4 \\
& Far FPR95$\downarrow$ & 64.1 & 51.2 & 67.0 & 93.1 & 93.1 & 91.5 & 92.1 \\
\addlinespace[2pt]
\textbf{LongReMix} & Accuracy$\uparrow$ & 96.3 & 95.1 & 93.8 & 79.9 & 94.7 & 77.9 & 75.5 \\
& Near AUROC$\uparrow$ & 79.5 & 80.2 & 76.8 & 73.7 & 77.5 & 77.1 & 76.2 \\
& Near FPR95$\downarrow$ & 63.0 & 51.9 & 58.9 & 82.5 & 72.1 & \underline{78.8} & 81.3 \\
& Far AUROC$\uparrow$ & 70.8 & 67.6 & 62.6 & 59.3 & 70.5 & 64.2 & \underline{72.5} \\
& Far FPR95$\downarrow$ & 64.9 & 58.6 & 62.5 & 88.0 & 68.6 & 90.0 & 85.7 \\
\addlinespace[2pt]
\textbf{L2B} & Accuracy$\uparrow$ & 92.1 & 88.4 & 71.4 & 50.6 & 91.9 & 68.8 & 58.4 \\
& Near AUROC$\uparrow$ & 76.0 & 72.3 & 67.5 & 61.6 & 75.3 & 71.1 & 68.4 \\
& Near FPR95$\downarrow$ & 62.9 & 73.2 & 89.6 & 91.2 & 68.1 & 86.9 & 88.0 \\
& Far AUROC$\uparrow$ & 74.7 & 75.5 & 69.9 & 60.2 & 78.9 & 69.9 & 60.3 \\
& Far FPR95$\downarrow$ & 61.3 & 64.9 & 86.6 & 83.9 & 66.4 & 87.7 & 93.7 \\
\addlinespace[2pt]
\textbf{PSSCL} & Accuracy$\uparrow$ & 96.4 & 95.6 & 93.7 & 92.9 & 93.9 & 77.6 & 77.0 \\
& Near AUROC$\uparrow$ & \textbf{91.3} & \textbf{90.7} & 83.3 & 65.6 & \textbf{90.6} & 70.5 & \textbf{77.4} \\
& Near FPR95$\downarrow$ & \textbf{35.8} & \textbf{35.8} & 41.7 & 89.6 & \textbf{39.7} & 87.9 & \textbf{76.5} \\
& Far AUROC$\uparrow$ & \textbf{93.4} & \underline{92.3} & \underline{87.8} & 64.2 & \underline{87.5} & \underline{66.1} & \underline{73.2} \\
& Far FPR95$\downarrow$ & \underline{24.0} & 26.4 & \underline{31.1} & 85.7 & \underline{33.8} & 79.5 & 73.5 \\
\addlinespace[2pt]
\textbf{UNICON} & Accuracy$\uparrow$ & 95.1 & 93.7 & 92.1 & 90.8 & 91.7 & \underline{78.9} & \underline{77.6} \\
& Near AUROC$\uparrow$ & 82.5 & 81.6 & 83.0 & \underline{84.1} & 83.1 & \underline{77.5} & 75.7 \\
& Near FPR95$\downarrow$ & 48.1 & 46.2 & 47.6 & 57.5 & 44.3 & 78.6 & 80.5 \\
& Far AUROC$\uparrow$ & \underline{84.7} & \textbf{92.9} & 83.9 & \textbf{92.8} & 82.5 & \textbf{76.2} & \textbf{76.1} \\
& Far FPR95$\downarrow$ & 38.4 & \underline{19.9} & 35.3 & \underline{32.0} & 33.9 & \textbf{67.9} & \textbf{71.3} \\
\addlinespace[2pt]
\textbf{RRL} & Accuracy$\uparrow$ & 95.9 & 93.9 & 83.4 & 87.6 & 93.7 & \textbf{79.5} & 74.6 \\
& Near AUROC$\uparrow$ & 54.2 & 48.3 & 39.3 & 49.9 & 46.2 & 52.5 & 47.8 \\
& Near FPR95$\downarrow$ & 92.5 & 98.6 & 97.0 & 96.0 & 95.4 & 92.3 & 96.1 \\
& Far AUROC$\uparrow$ & 66.3 & 51.1 & 30.8 & 61.3 & 41.7 & 55.7 & 56.6 \\
& Far FPR95$\downarrow$ & 84.9 & 98.8 & 99.1 & 85.6 & 98.0 & 97.5 & 96.4 \\
\addlinespace[2pt]
\textbf{ProMix} & Accuracy$\uparrow$ & \textbf{97.6} & \textbf{97.3} & \textbf{94.0} & 92.8 & 93.9 & \textbf{82.6} & \textbf{79.3} \\
& Near AUROC$\uparrow$ & \underline{88.1} & \underline{87.2} & \textbf{87.4} & 83.0 & \underline{85.5} & 73.3 & 76.1 \\
& Near FPR95$\downarrow$ & \underline{37.5} & \underline{38.4} & \underline{41.6} & \underline{51.2} & \underline{40.6} & 80.2 & 81.5 \\
& Far AUROC$\uparrow$ & 76.6 & 82.0 & 79.4 & 82.0 & 81.2 & 51.9 & 63.9 \\
& Far FPR95$\downarrow$ & 45.6 & 36.0 & 40.5 & 41.7 & 39.2 & 93.8 & 90.9 \\
\bottomrule
\end{tabular}
}
\end{table*}

\section{Accuracy and Open-World Reliability Decouple Under Stress}
\label{sec:alignment}

\begin{figure*}[!htbp]
\centering
\setlength{\tabcolsep}{2pt}
\begin{tabular}{@{}ccc@{}}
\includegraphics[width=0.32\linewidth]{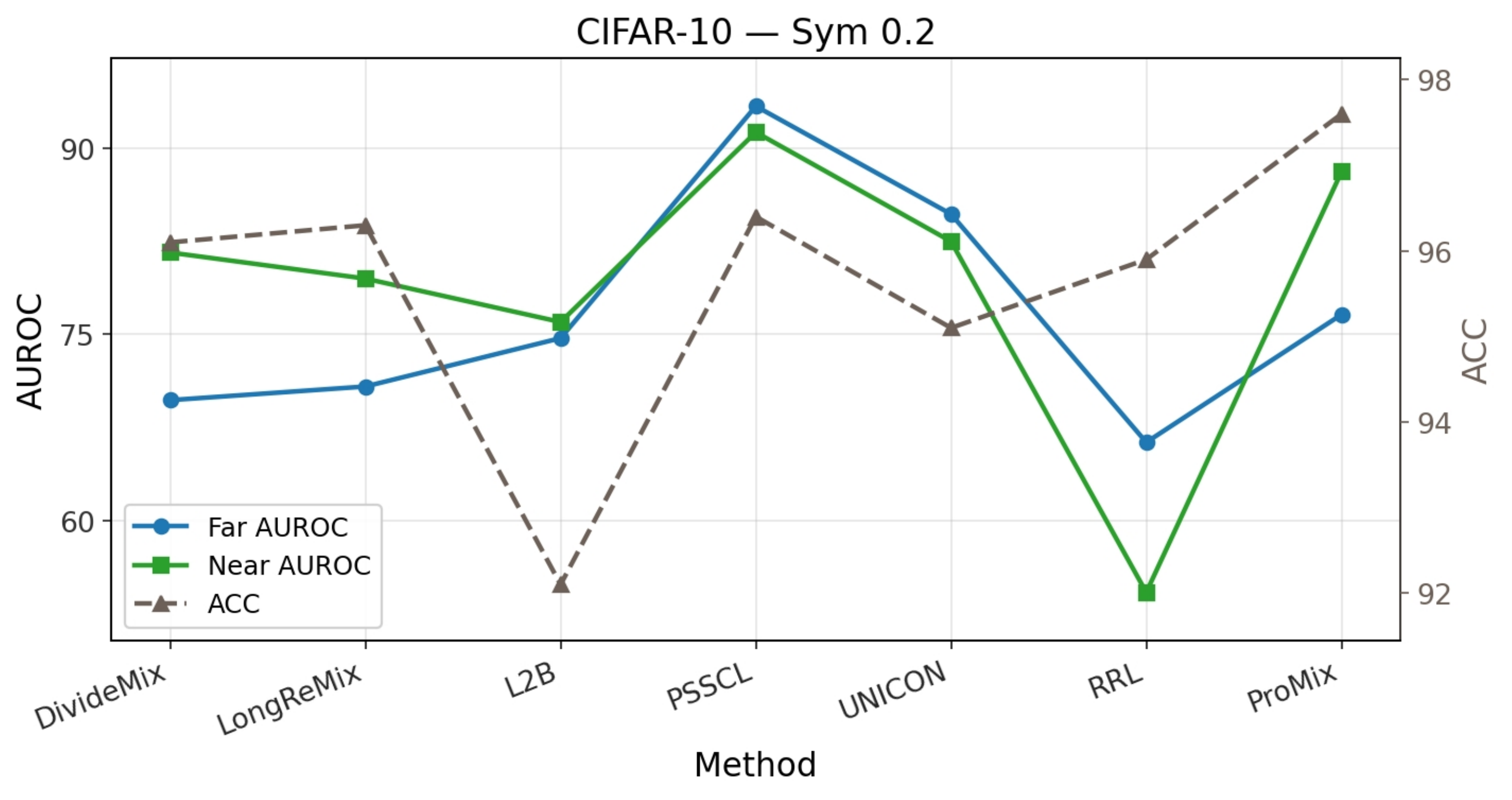} &
\includegraphics[width=0.32\linewidth]{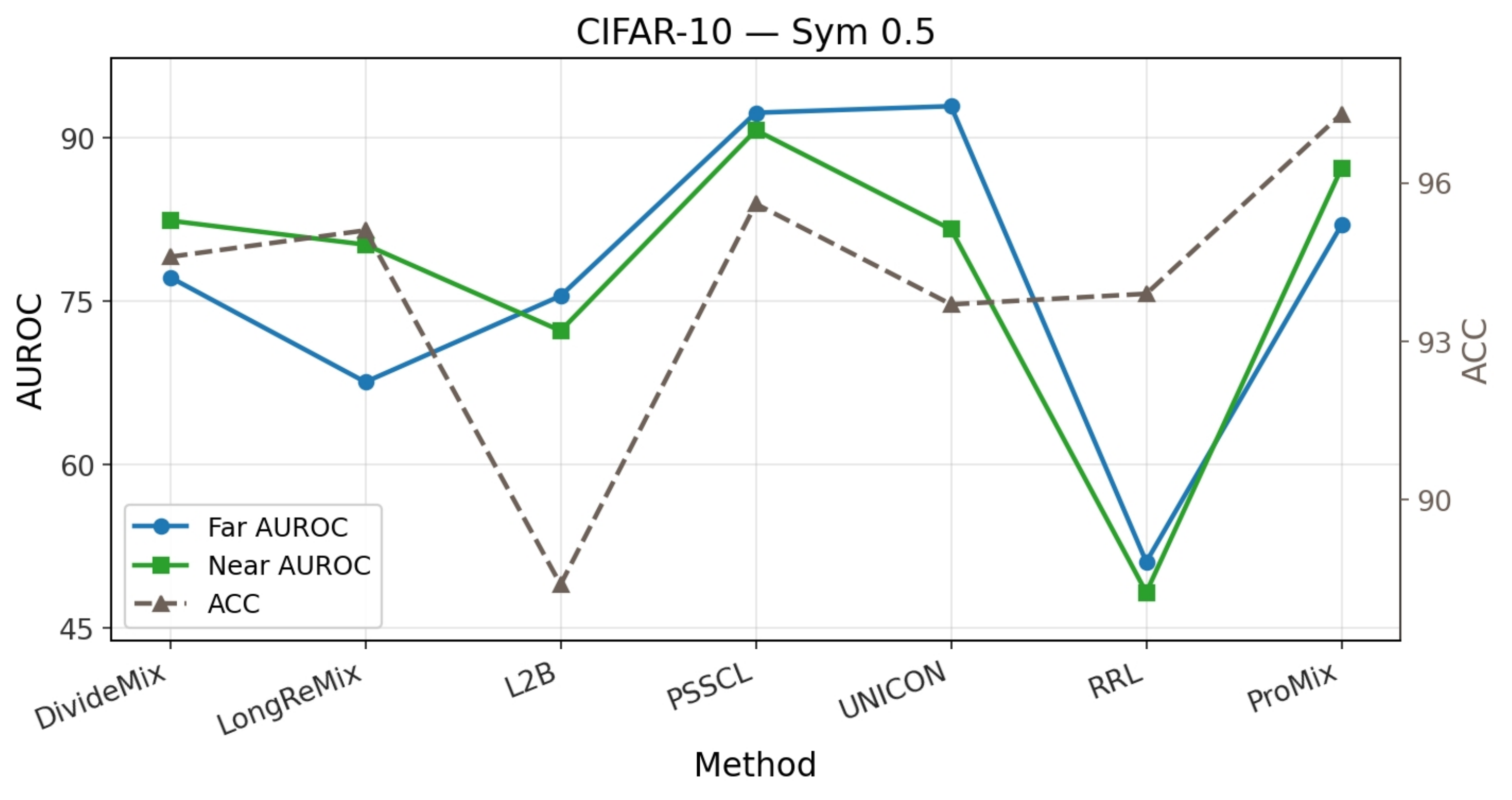} &
\includegraphics[width=0.32\linewidth]{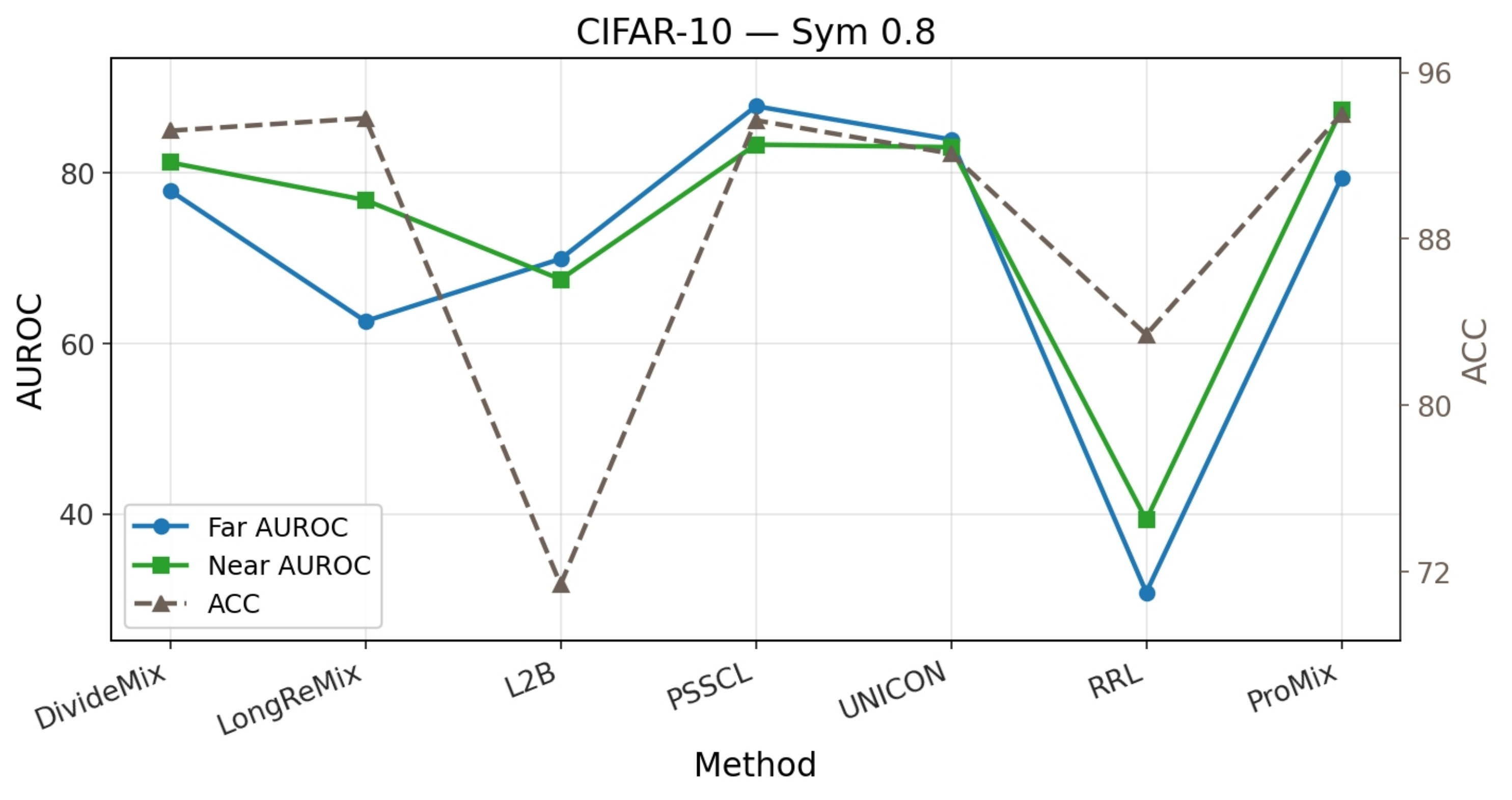} \\
\includegraphics[width=0.32\linewidth]{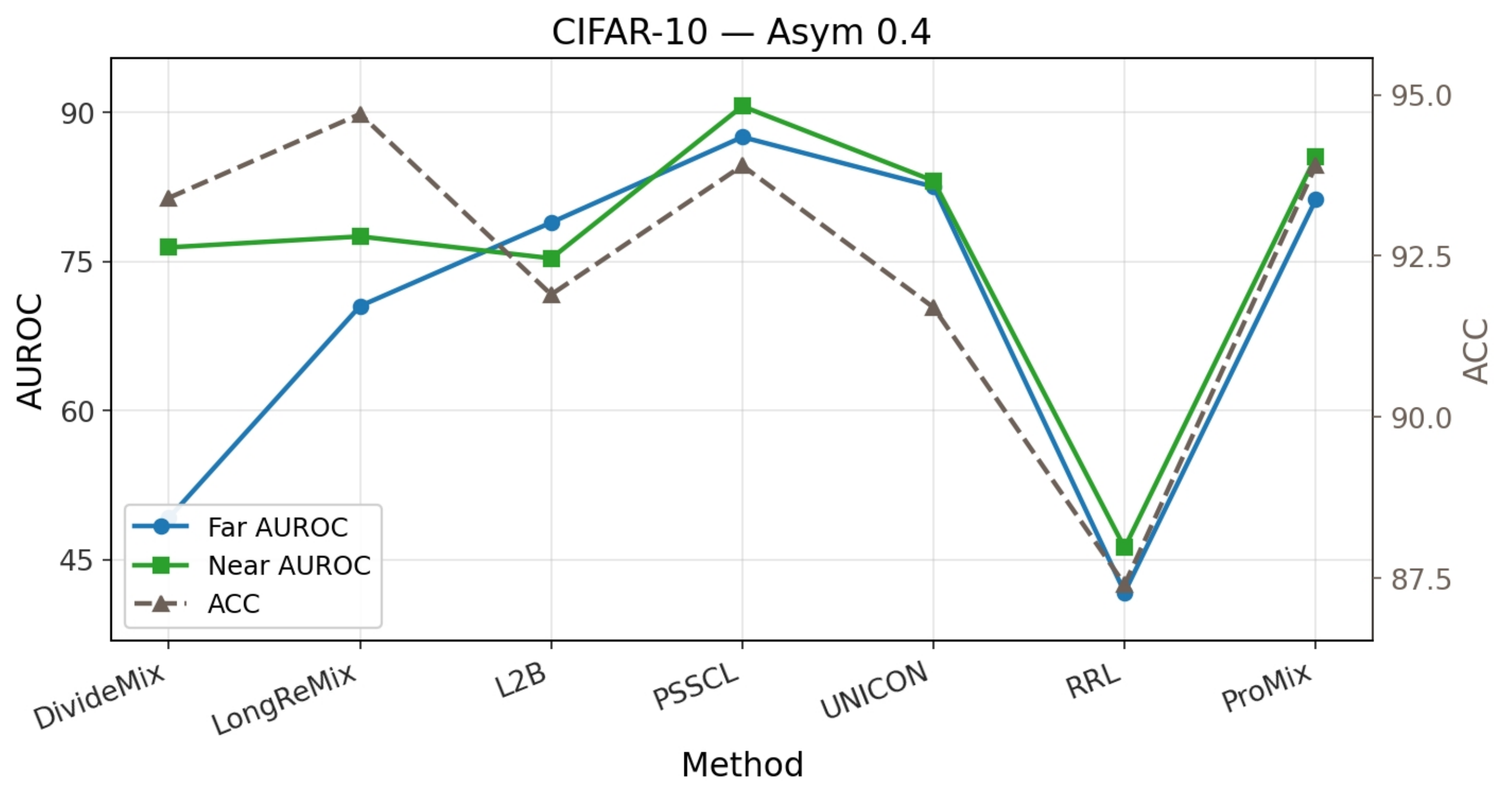} &
\includegraphics[width=0.32\linewidth]{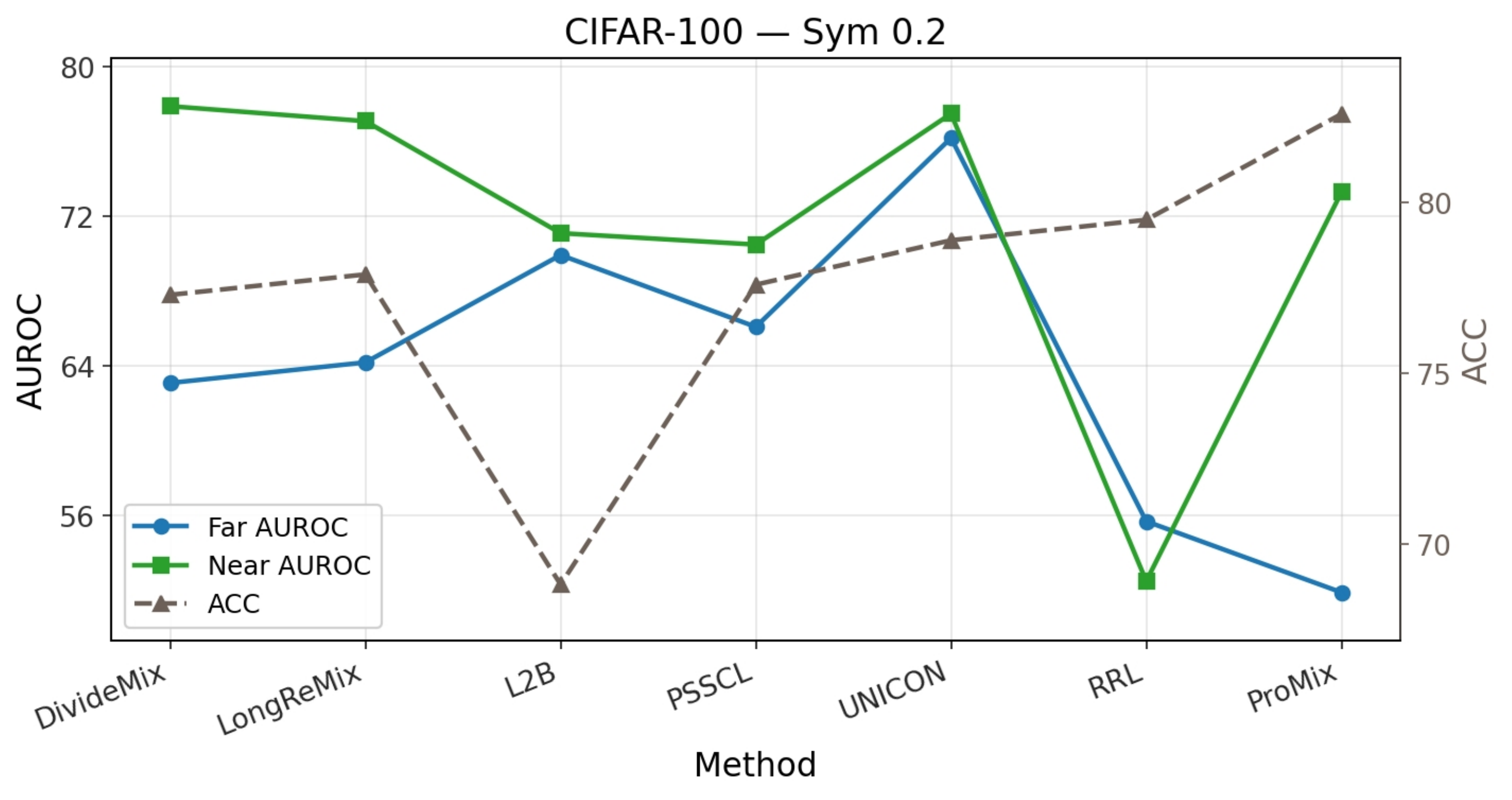} &
\includegraphics[width=0.32\linewidth]{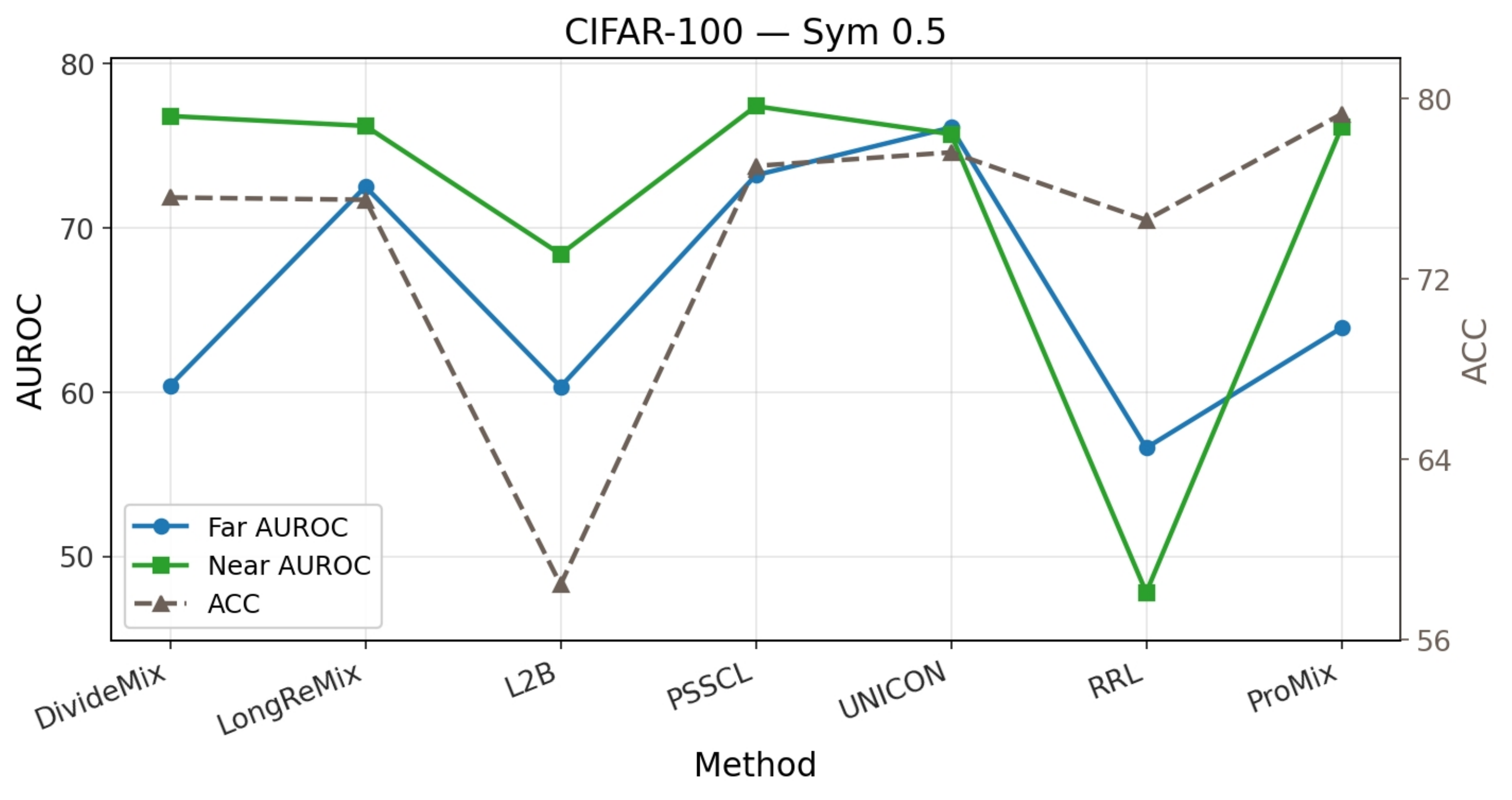} \\
\end{tabular}
\caption{\textbf{Per-noise ACC--OOD profiles corresponding to Table~\ref{tab:main-benchmark}.}
Each panel groups methods within one dataset/noise setting and plots closed-set ACC together with near- and far-OOD AUROC, making visible that higher closed-set accuracy does not consistently imply better OOD reliability under the frozen-checkpoint contract.}
\label{fig:table1-strength-by-noise}
\end{figure*}

\noindent
A common accuracy-centric assumption is that a stronger noisy-label classifier should also yield a more reliable OOD detector under the same frozen-network contract.
As shown in Table~\ref{tab:main-benchmark}, the CIFAR-100 symmetric-$0.5$ setting provides a useful stress point: several learners occupy a narrow accuracy band yet spread widely across near- and far-OOD AUROC.
Representative settings with PreAct-ResNet-50 under the same protocol are omitted in this arXiv preprint version.
Figure~\ref{fig:table1-strength-by-noise} visualizes the same table entries by noise setting, making the row-wise ACC/OOD decoupling easier to compare.

\paragraph{What the benchmark shows.}
Methods that are close in ACC can differ widely in OOD reliability, and the best-ACC method is often not the best OOD method.

On CIFAR-100 sym\,$0.5$, the comparable high-ACC baselines lie in a narrow ACC band, yet near-OOD AUROC ranges from $47.8$ (RRL) to $77.4$ (PSSCL), and far-OOD AUROC ranges from $56.6$ (RRL) to $76.1$ (UNICON).
L2B is outside this accuracy band in this setting; we report it for completeness but do not use that cell as evidence for the high-ACC decoupling claim.
These separations show why closed-set accuracy alone is an incomplete LNL benchmark.
The RN50 ablation follows the same pattern qualitatively: widening the backbone does not by itself guarantee that ACC and OOD reliability align.

\paragraph{Near- and Far-OOD Represent Distinct Failure Axes.}
Near- and far-OOD often agree in broad trends, but their leaders can differ.
The far-OOD suite stress-tests coarse separation from external images, while the near-OOD contract stress-tests whether fine-grained semantic neighborhoods remain disentangled from the ID-wrong stratum.
Section~\ref{sec:pathology} unifies both under uncertainty collapse.

\section{Real-Noise Validation}
\label{sec:real-noise}

To test whether the observed ACC--OOD decoupling is specific to synthetic noise generation, we extend the same protocol to real-world annotation noise on Food-101N, Animal-10N, and CIFAR-N.
Table~\ref{tab:real-noise-front} shows mixed real-noise behavior: Food-101N aligns ACC and OOD metrics for RRL, while Animal-10N and CIFAR-N include cases where closed-set accuracy and far-OOD reliability separate.

\begin{table*}[!htbp]
\centering
\caption{\textbf{Real-noise validation.}
Food-101N and Animal-10N report ACC, far-OOD AUROC, and far-OOD FPR95; CIFAR-N reports ACC and far-OOD AUROC for each split.
Far-OOD scores follow the fixed external-suite protocol from Section~\ref{sec:setup}.
Panels \textbf{(a)} and \textbf{(b)} are placed side by side; panel \textbf{(c)} is aligned below them.
Bold marks the best value per column and underline marks the runner-up when unique.}
\label{tab:real-noise-front}
\label{tab:large-scale-far}
\label{tab:real-noise}
\scriptsize
\setlength{\tabcolsep}{3.2pt}
\renewcommand{\arraystretch}{1.06}
\begin{minipage}[t]{0.92\textwidth}
\centering
\begin{minipage}[t]{0.48\linewidth}
\centering
\textbf{(a) Food-101N}\\[0.35em]
\resizebox{\linewidth}{!}{%
\begin{tabular}{@{}lccc@{}}
\toprule
\textbf{Method} & ACC$\uparrow$ & AUROC$\uparrow$ & FPR95$\downarrow$ \\
\midrule
\textbf{DivideMix} & 70.2 & 95.1 & 22.4 \\
\textbf{L2B}       & 69.8 & \underline{97.2} & \textbf{10.4} \\
\textbf{LongReMix} & 62.9 & 82.8 & 60.5 \\
\textbf{ProMix}    & 56.9 & 94.2 & 17.9 \\
\textbf{PSSCL}     & 63.1 & 83.3 & 65.7 \\
\textbf{RRL}       & \textbf{85.0} & \textbf{97.7} & \textbf{10.4} \\
\textbf{UNICON}    & \underline{76.5} & 94.8 & 28.1 \\
\bottomrule
\end{tabular}%
}
\end{minipage}\hspace{0.035\linewidth}
\begin{minipage}[t]{0.48\linewidth}
\centering
\textbf{(b) Animal-10N}\\[0.35em]
\resizebox{\linewidth}{!}{%
\begin{tabular}{@{}lccc@{}}
\toprule
\textbf{Method} & ACC$\uparrow$ & AUROC$\uparrow$ & FPR95$\downarrow$ \\
\midrule
\textbf{DivideMix} & 76.7 & 77.2 & 85.9 \\
\textbf{L2B}       & \underline{81.4} & 82.6 & 74.2 \\
\textbf{LongReMix} & 71.6 & 81.9 & 66.7 \\
\textbf{ProMix}    & \textbf{84.1} & 83.0 & \underline{62.4} \\
\textbf{PSSCL}     & 72.6 & 75.0 & 74.9 \\
\textbf{RRL}       & 73.9 & \underline{85.7} & 67.2 \\
\textbf{UNICON}    & 71.7 & \textbf{87.2} & \textbf{55.6} \\
\bottomrule
\end{tabular}%
}
\end{minipage}
\end{minipage}

\vspace{0.6em}

\begin{minipage}[t]{0.92\textwidth}
\centering
\textbf{(c) CIFAR-N}\\[0.35em]
\resizebox{\linewidth}{!}{%
\begin{tabular}{@{}lcccccccc@{}}
\toprule
\textbf{Method}
& \multicolumn{2}{c}{C10N-A}
& \multicolumn{2}{c}{C10N-R1}
& \multicolumn{2}{c}{C10N-W}
& \multicolumn{2}{c}{C100N} \\
\cmidrule(lr){2-3}\cmidrule(lr){4-5}\cmidrule(lr){6-7}\cmidrule(l){8-9}
& ACC$\uparrow$ & AUROC$\uparrow$ & ACC$\uparrow$ & AUROC$\uparrow$ & ACC$\uparrow$ & AUROC$\uparrow$ & ACC$\uparrow$ & AUROC$\uparrow$ \\
\midrule
\textbf{LongReMix} & 87.4 & 78.0 & 90.3 & 78.0 & 92.5 & 73.6 & 70.7 & 68.0 \\
\textbf{L2B}       & 89.8 & \textbf{90.8} & 88.2 & \textbf{90.8} & 83.1 & \underline{87.6} & 58.9 & \textbf{79.2} \\
\textbf{ProMix}    & \textbf{97.1} & \underline{88.1} & \textbf{96.9} & \underline{88.1} & \textbf{94.2} & 85.9 & \underline{71.7} & 69.8 \\
\textbf{PSSCL}     & 93.0 & 82.6 & 93.6 & 82.6 & 91.5 & 84.0 & \textbf{74.2} & \underline{78.0} \\
\textbf{RRL}       & \underline{95.4} & 63.9 & \underline{94.6} & 48.4 & 85.7 & 42.9 & 61.3 & 54.5 \\
\textbf{UNICON}    & 93.9 & 86.4 & 94.0 & 86.4 & \underline{93.7} & \textbf{93.7} & 65.7 & 75.5 \\
\bottomrule
\end{tabular}%
}
\end{minipage}
\end{table*}

The real-noise results are mixed rather than uniformly inverted.
Food-101N is an aligned case: RRL has the best ACC, AUROC, and FPR95.
Animal-10N and CIFAR-N still show separations between closed-set accuracy and far-OOD reliability, so Table~\ref{tab:real-noise-front} supports reporting both quantities under real annotation noise.

\providecommand{\Cscore}{\mathcal{C}_{\mathrm{score}}}

\section{Core Pathology: Uncertainty Collapse}
\label{sec:pathology}

We trace the ACC--OOD gap to score geometry.
Label noise can create overlap between misclassified ID (ID-wrong) and OOD samples~\cite{noisyelephant}, but the overlap is not a uniform shift of all errors.
It concentrates in \emph{uncertainty collapse}: a low-confidence ID-wrong subpopulation overlaps OOD score support, making novelty detection confound true OOD inputs with unresolved ID mistakes.
Corrupted labels can therefore leave a persistent ambiguous stratum even when the final classifier appears accurate on clean ID test labels.
Table~\ref{tab:five-group} localizes this effect; Section~\ref{sec:intervention} later tests whether a repair changes the same structure rather than only headline AUROC.

\paragraph{Accuracy gains do not imply structural recovery.}
The moderate aggregate ACC--AUROC correlation ($\rho \approx 0.71$) masks a key limitation: closed-set gains do not guarantee recovery of a stable discriminative geometry.
Robust objectives can reduce error while leaving low-confidence ID mistakes aligned with OOD scores, weakening abstention.
This matters for post-hoc OOD detection because the detector inherits the frozen classifier's score and feature geometry rather than an independently trained rejection boundary.

\paragraph{Formal scores and stratified AUROC.}
Let $f_\theta:\mathcal{X}\!\to\!\mathbb{R}^K$ be the frozen classifier with logits $z(x)=f_\theta(x)$.
We collect MSP confidence, Energy~\cite{liu2020energy}, and the misclassified-ID index set in one display (fixed $T{=}1$ throughout):
\begin{align}
  c_{\mathrm{MSP}}(x) &= \max_{k\in\{1,\ldots,K\}} \softmax(z(x))_k ,
  \label{eq:msp-conf}
  \\
  E(x) &= -T\log\sum_{k=1}^K \exp\!\bigl(z_k(x)/T\bigr),
  \label{eq:energy-eval}
  \\
  \mathcal{S}_{\mathrm{wrong}}
    &= \bigl\{x\in\mathcal{D}_{\mathrm{ID}}^{\mathrm{test}}:\operatorname{argmax}_{k} z_k(x)\neq y(x)\bigr\}.
  \label{eq:Swrong}
\end{align}
Here $\mathcal{D}_{\mathrm{ID}}^{\mathrm{test}}$ uses clean test labels, $\mathcal{D}_{\mathrm{OOD}}$ is pooled OOD, and larger $E(x)$ means stronger OOD evidence.
For a scalar score $s$, AUROC is computed after converting all scores to the common OOD-larger direction.
For \emph{ID-wrong vs.\ OOD}, we report AUROC between negatives $\mathcal{S}_{\mathrm{wrong}}$ and positives $\mathcal{D}_{\mathrm{OOD}}$.
The five-group breakdown splits ID-correct and ID-wrong samples at the MSP median, yielding high/low-confidence strata for each correctness group.

\paragraph{Low-confidence ID errors act as OOD impostors.}
The collapse is driven mainly by low-confidence ID-wrong samples.
They carry $93.2\%$ of ID-wrong mass in the pooled aggregate and remain dominant in most Table~\ref{tab:five-group} regimes.
High-confidence errors are often more separable, while low-confidence errors occupy the score region used to reject anomalies.
Thus a threshold tuned to reject OOD inputs can also remove a substantial fraction of unresolved ID mistakes.


\paragraph{Two layers of the same failure.}
Far-OOD degradation reflects loss of coarse separation from external inputs; near-OOD degradation reflects finer semantic interleaving with adjacent datasets.
An intervention might therefore widen the far-OOD margin while leaving near-OOD harder to resolve.
This distinction is useful later when VMR improves far-OOD reliability more consistently than near-OOD reliability.

\paragraph{Score distributions expose relative compression and gap shrinkage.}
Figure~\ref{fig:energy-dist-grid} and Table~\ref{tab:five-group} summarize the high-stress score geometry on the same Energy axis used by the detector.
The density grid uses RRL as a representative stress case for distributional inspection, not as a method-specific claim.
Across regimes, the recurring pattern is a low-confidence basin in which ID-wrong mass enters the OOD Energy band, while the ID-correct anchor remains better separated.
ECE~\cite{guo2017calibration} and Energy AUROC~\cite{liu2020energy} correlate negatively across evaluation configurations ($\rho \approx -0.8$), indicating that miscalibration often co-occurs with score-space failure.
The conclusion is not Energy-specific: method rankings are nearly unchanged across MSP, MaxLogit, margin, entropy, and Energy, while detector controls do not outperform Energy in representative probes.
Because ratio-style collapse summaries depend on score scaling, Section~\ref{sec:intervention} uses paired before/after diagnostics under the same host learner.
These paired diagnostics are used only for within-host repair analysis, not for cross-method leaderboard ranking.

\paragraph{What the taxonomy rules out.}
The five-group view also argues against a simpler explanation in which OOD failure is caused only by a globally weak detector.
ID-correct samples remain comparatively anchored, and high-confidence mistakes are not the dominant overlap source.
The problem is concentrated in a specific low-confidence ID-error reservoir, which is why aggregate ID-vs.-OOD AUROC can hide the mechanism.

\subsection{Clean-vs-noise comparisons implicate training pressure}
\label{sec:noise-pressure}

To separate noise pressure from inherent image ambiguity, we compare noise-trained checkpoints with clean-label baselines.
Switching from clean to noisy training can drive ID-wrong vs.\ OOD AUROC from $\approx\!71\%$ to $\approx\!50\%$ for DivideMix even as the ID-error count decreases.
The relevant change is therefore not only how many ID errors remain, but where those errors sit relative to the OOD score support.

\begingroup
\centering
\footnotesize
\setlength{\tabcolsep}{2.4pt}
\medskip
\captionof{table}{\textbf{Five-group taxonomy} (Energy score). Columns are synthetic-noise regimes; within each regime, AU is ID-subgroup vs.\ pooled far-OOD AUROC (five datasets) and m\% is mass on the ID test under the MSP median split.
The table decomposes fixed benchmark checkpoints after training; subgroup labels and AU values are diagnostic outputs, not checkpoint-selection criteria.
Host methods may differ across columns, so the table is not a cross-method leaderboard.
High-confidence ID-correct samples do not require an OOD-separability calculation for this collapse diagnostic, so their AU values are intentionally not computed and shown as \texttt{---}.}
\label{tab:five-group}
\vspace{0.35em}
\resizebox{\textwidth}{!}{%
\begin{tabular}{@{}l cc cc cc cc cc cc cc cc@{}}
\toprule
 & \multicolumn{2}{c}{C10 sym\,0.2} & \multicolumn{2}{c}{C10 sym\,0.5} & \multicolumn{2}{c}{C10 sym\,0.8} & \multicolumn{2}{c}{C10 sym\,0.9} & \multicolumn{2}{c}{C10 asym\,0.4} & \multicolumn{2}{c}{C100 sym\,0.2} & \multicolumn{2}{c}{C100 sym\,0.5} \\
\cmidrule(lr){2-3}\cmidrule(lr){4-5}\cmidrule(lr){6-7}\cmidrule(lr){8-9}\cmidrule(lr){10-11}\cmidrule(lr){12-13}\cmidrule(lr){14-15}
Group & AU$\uparrow$ & m\% & AU$\uparrow$ & m\% & AU$\uparrow$ & m\% & AU$\uparrow$ & m\% & AU$\uparrow$ & m\% & AU$\uparrow$ & m\% & AU$\uparrow$ & m\% \\
\midrule
ID-correct high-conf
  & --- & 49.8 & --- & 49.4 & --- & 45.4 & --- & 29.3 & --- & 47.7 & --- & 34.4 & --- & 25.4 \\
ID-correct low-conf
  & 0.8 & 46.2 & 0.7 & 44.3 & 0.6 & 26.1 & 0.4 & 14.8 & 0.7 & 35.7 & 0.6 & 13.3 & 0.3 & 7.0 \\
ID-wrong high-conf
  & 0.9 & 0.2 & 0.9 & 0.7 & 0.9 & 4.6 & 0.8 & 20.7 & 0.8 & 2.3 & 0.7 & 15.6 & 0.5 & 24.6 \\
\textbf{ID-wrong low-conf}
  & \textbf{0.5} & 3.9 & \textbf{0.5} & 5.7 & \textbf{0.5} & 23.9 & \textbf{0.3} & 35.2 & \textbf{0.5} & 14.3 & \textbf{0.4} & 36.8 & \textbf{0.2} & 43.0 \\
\bottomrule
\end{tabular}%
}
\endgroup
\smallskip

\paragraph{Representation geometry: numbers and mechanism.}
The collapse also appears in penultimate geometry (C100 sym\,$0.5$): participation ratios are $50$--$69$ for ID-correct, \textbf{$23$--$30$} for ID-wrong, and $35$--$42$ for OOD, with MLE intrinsic dimensions $12.6$, \textbf{$8.8$}, and $10.8$, respectively.
ID-wrong is therefore lower-dimensional and closer to OOD than to the anchor manifold.
Direction-based probes dominate ID-correct/OOD separation, yet ID-wrong aligns with OOD in angular drift ($\cos \approx 0.84$), indicating a systematic fragile region.
Misclassified IDs are also shared across methods at this noise level, consistent with a recurring fragile subpopulation rather than isolated implementation artifacts.

\smallskip\noindent
Figure~\ref{fig:energy-dist-grid} plots kernel-smoothed Energy densities for RRL on ID-correct, ID-wrong, and pooled far-OOD tests (shared top-left legend; seven synthetic-noise settings), rendering the low-confidence overlap on the same Energy axis.

\begingroup
\centering
\normalsize
\setlength{\arrayrulewidth}{0pt}
{\setlength{\tabcolsep}{0pt}%
\setlength{\energytilew}{\dimexpr\linewidth/4\relax}%
\begin{tabular}{@{}cccc@{}}
\includegraphics[width=\energytilew,trim=2 2 2 2,clip]{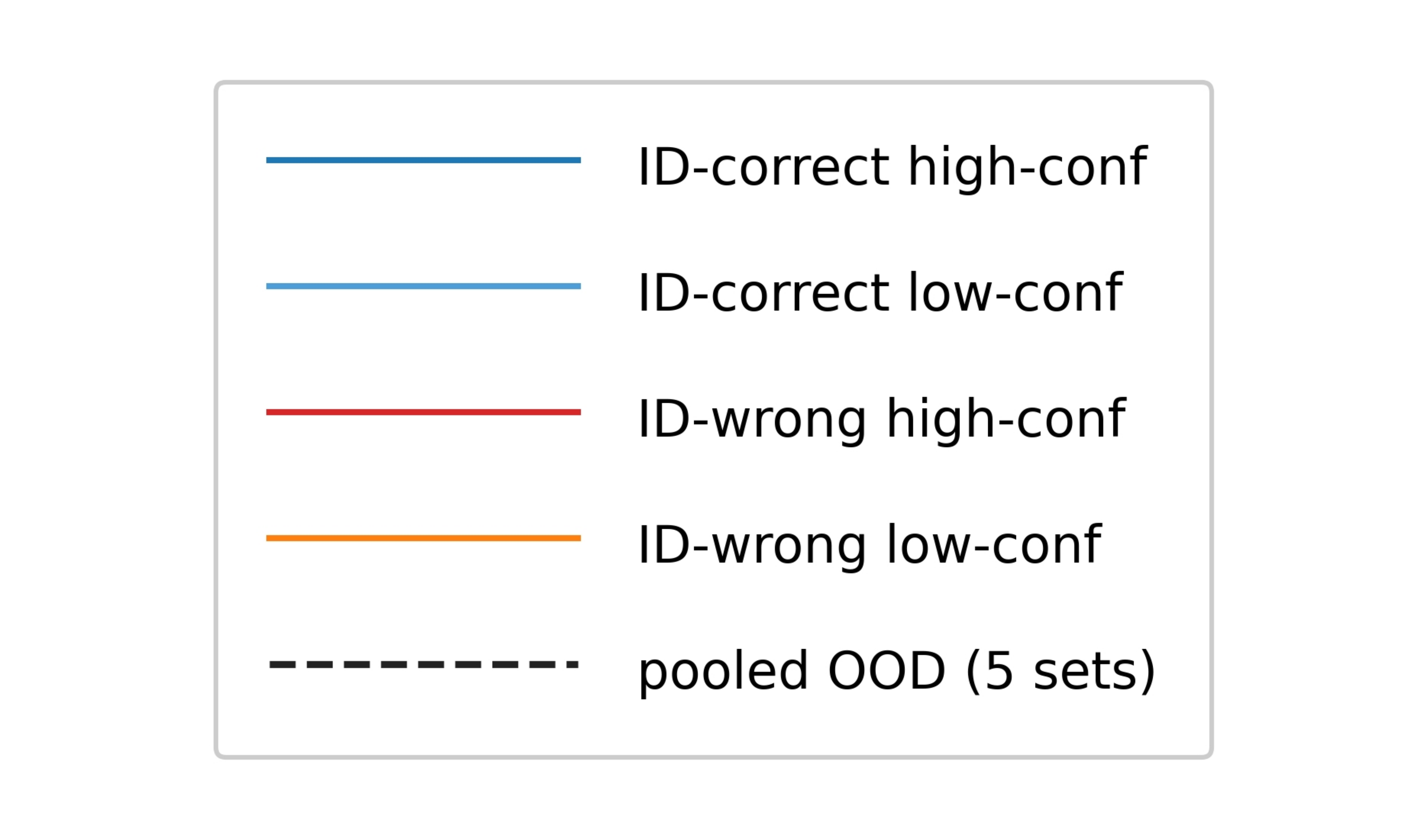} &
\includegraphics[width=\energytilew,trim=2 2 2 2,clip]{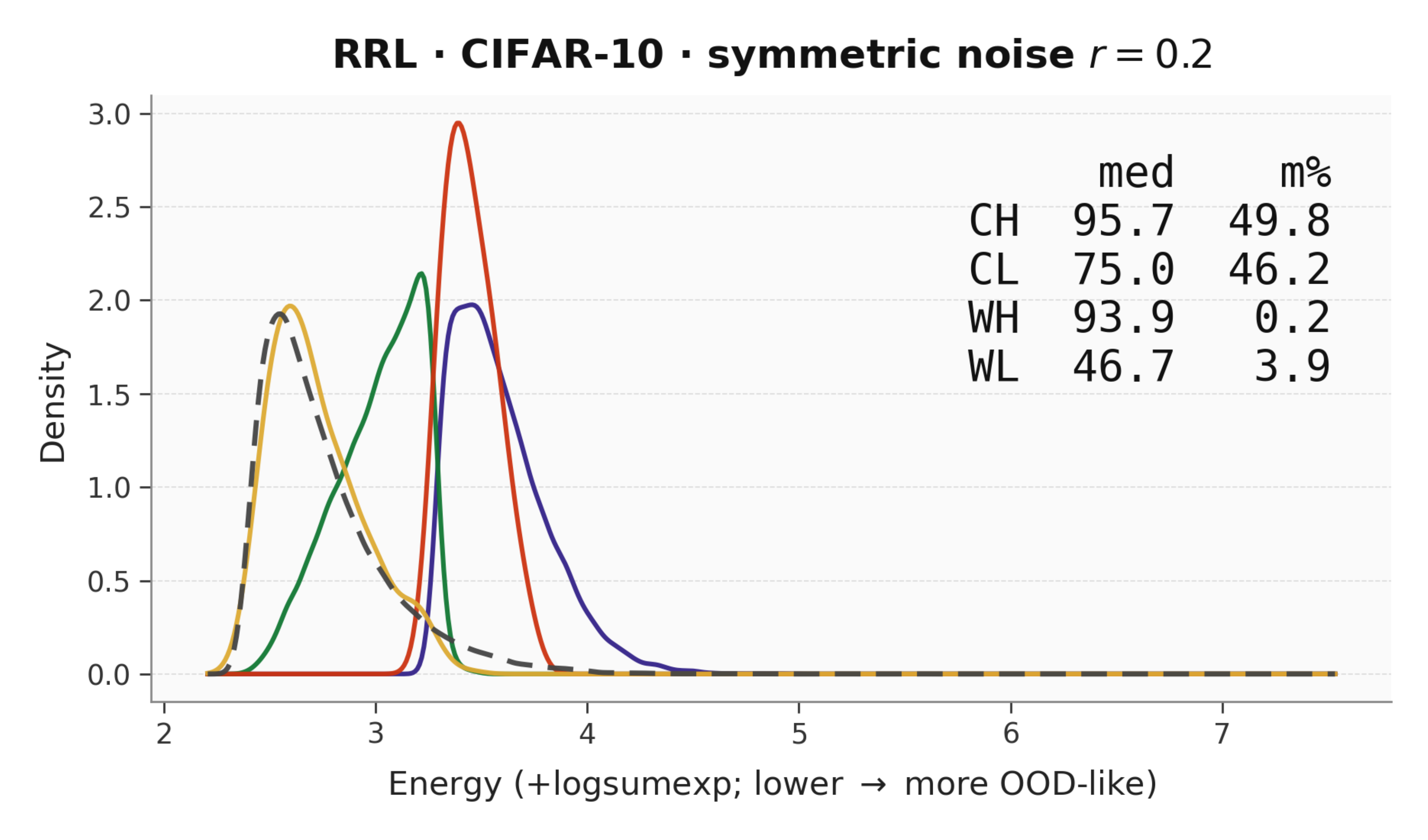} &
\includegraphics[width=\energytilew,trim=2 2 2 2,clip]{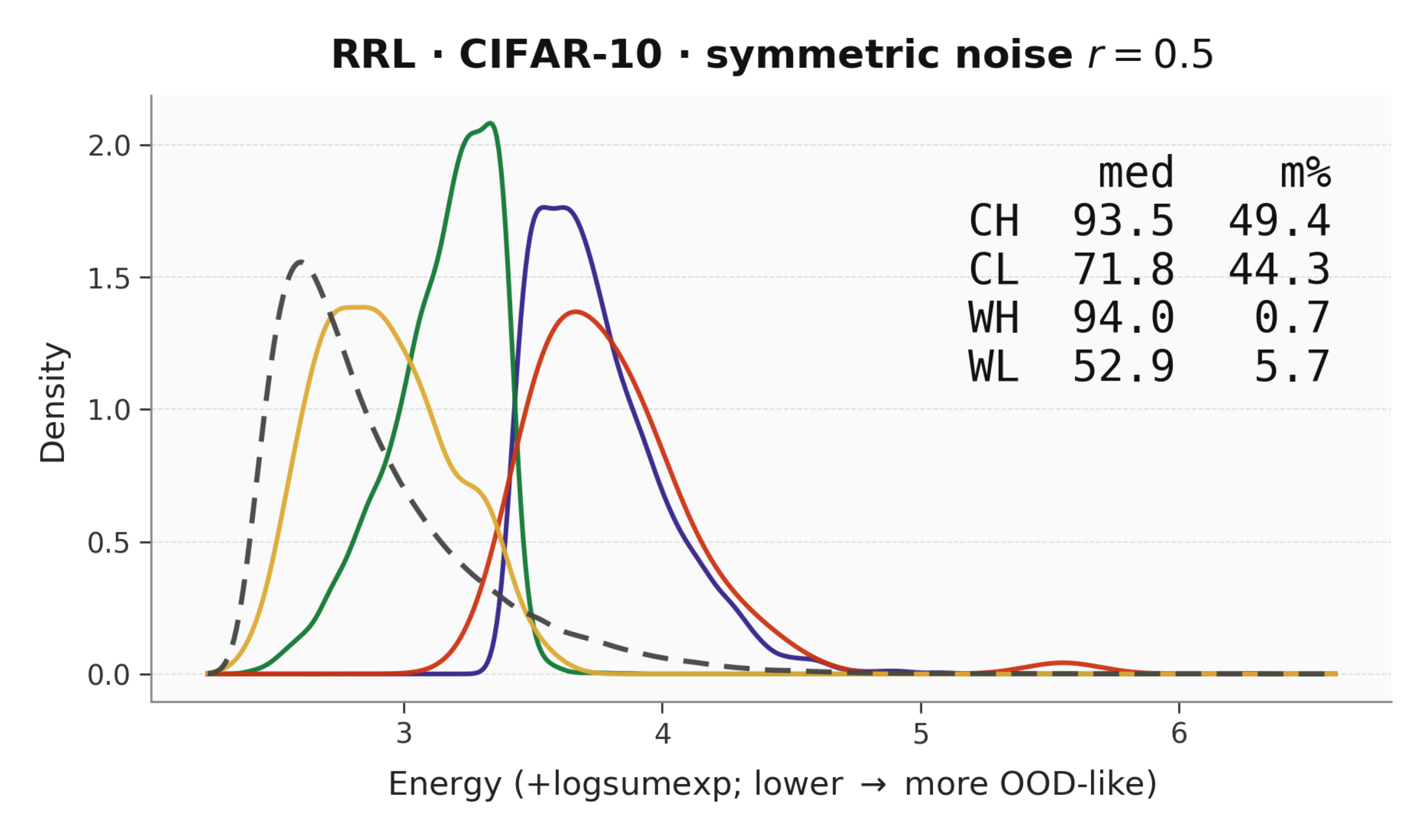} &
\includegraphics[width=\energytilew,trim=2 2 2 2,clip]{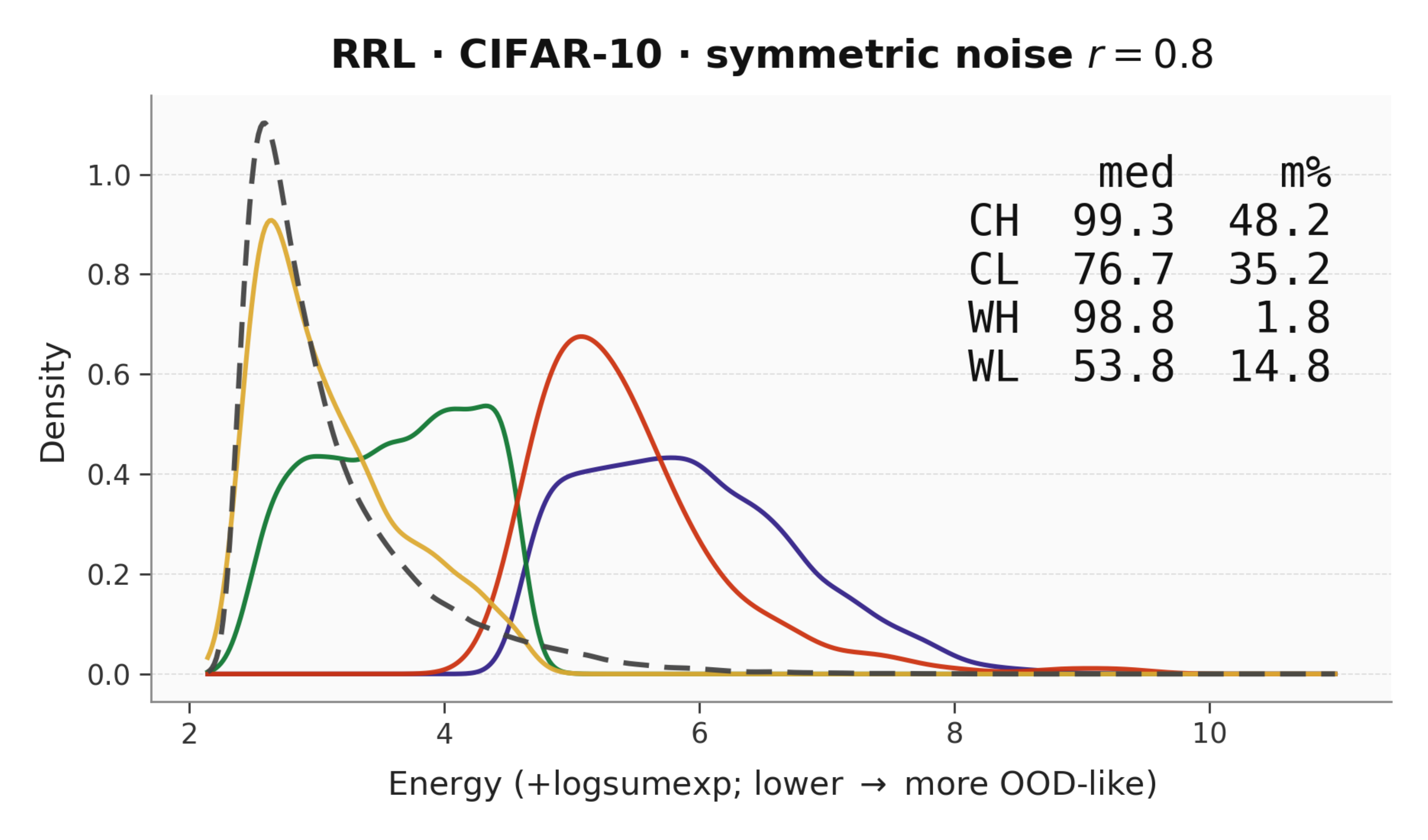} \\
\includegraphics[width=\energytilew,trim=2 2 2 2,clip]{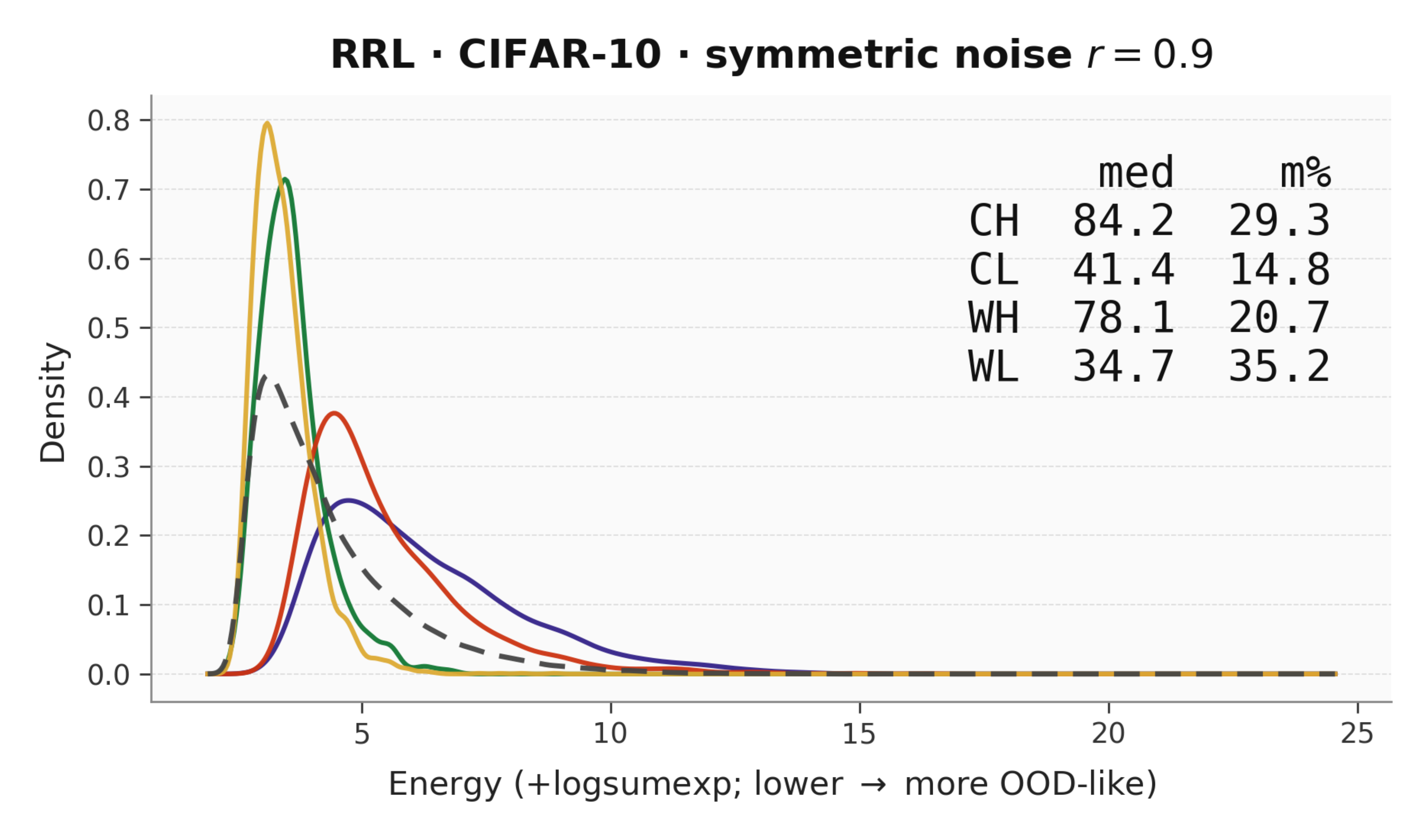} &
\includegraphics[width=\energytilew,trim=2 2 2 2,clip]{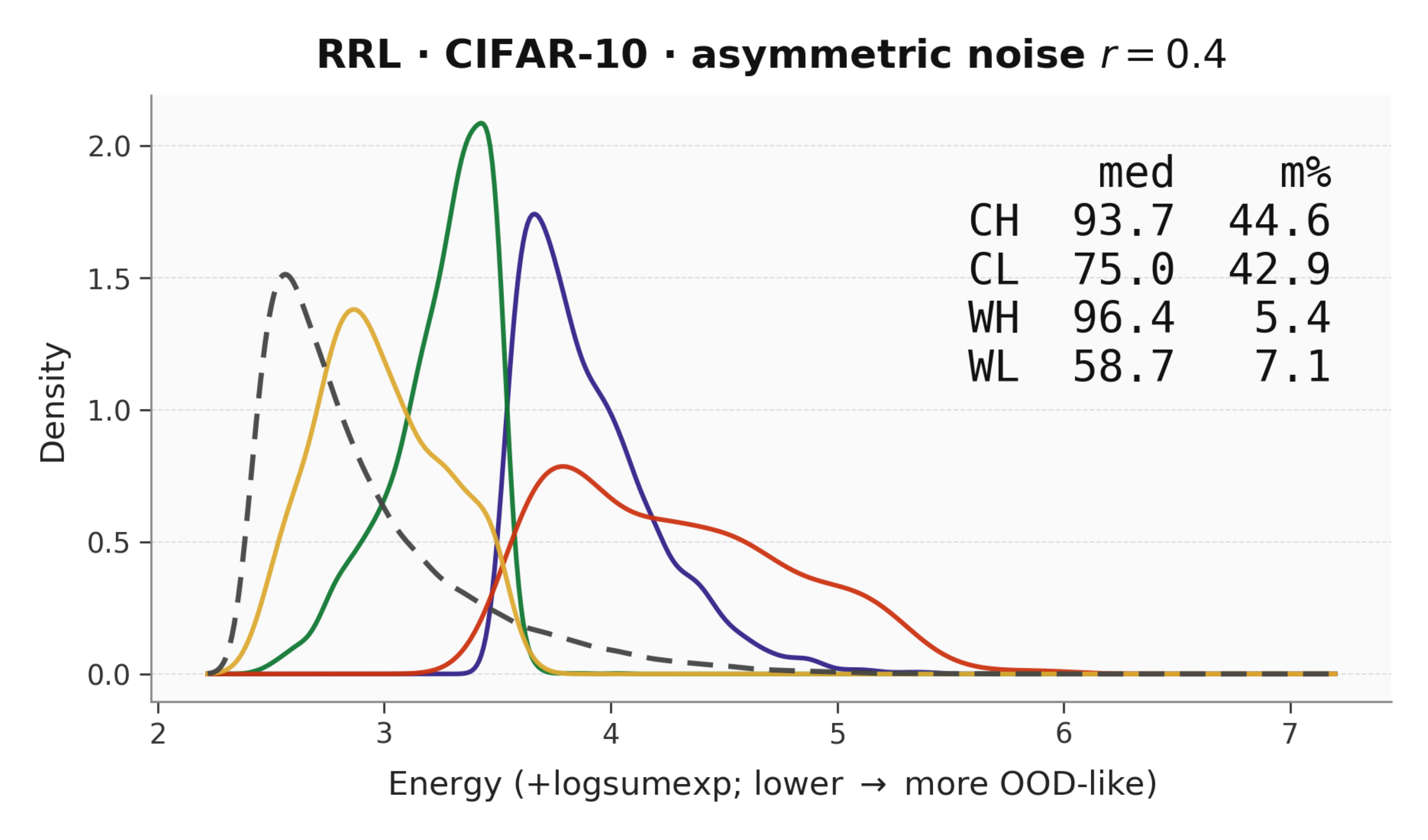} &
\includegraphics[width=\energytilew,trim=2 2 2 2,clip]{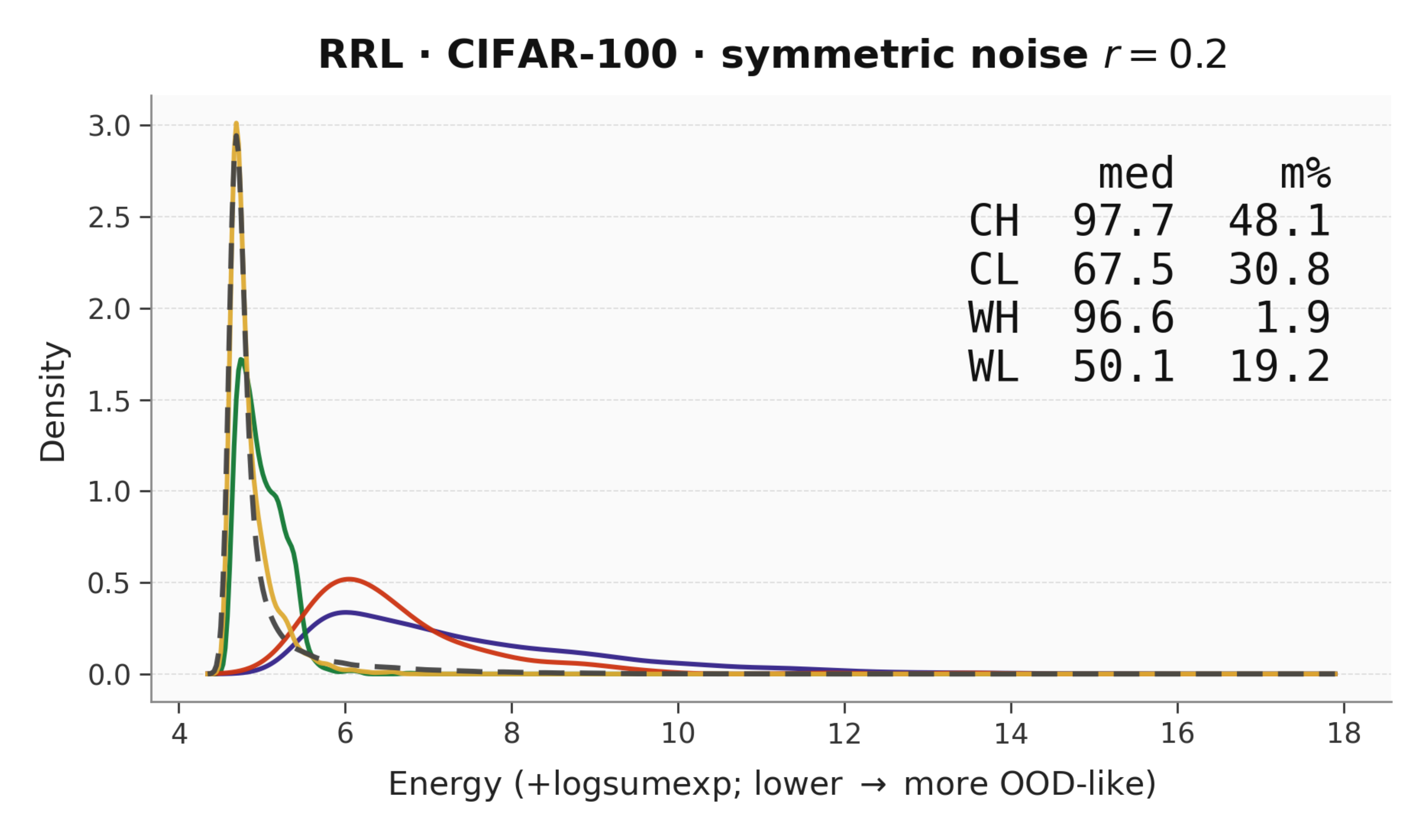} &
\includegraphics[width=\energytilew,trim=2 2 2 2,clip]{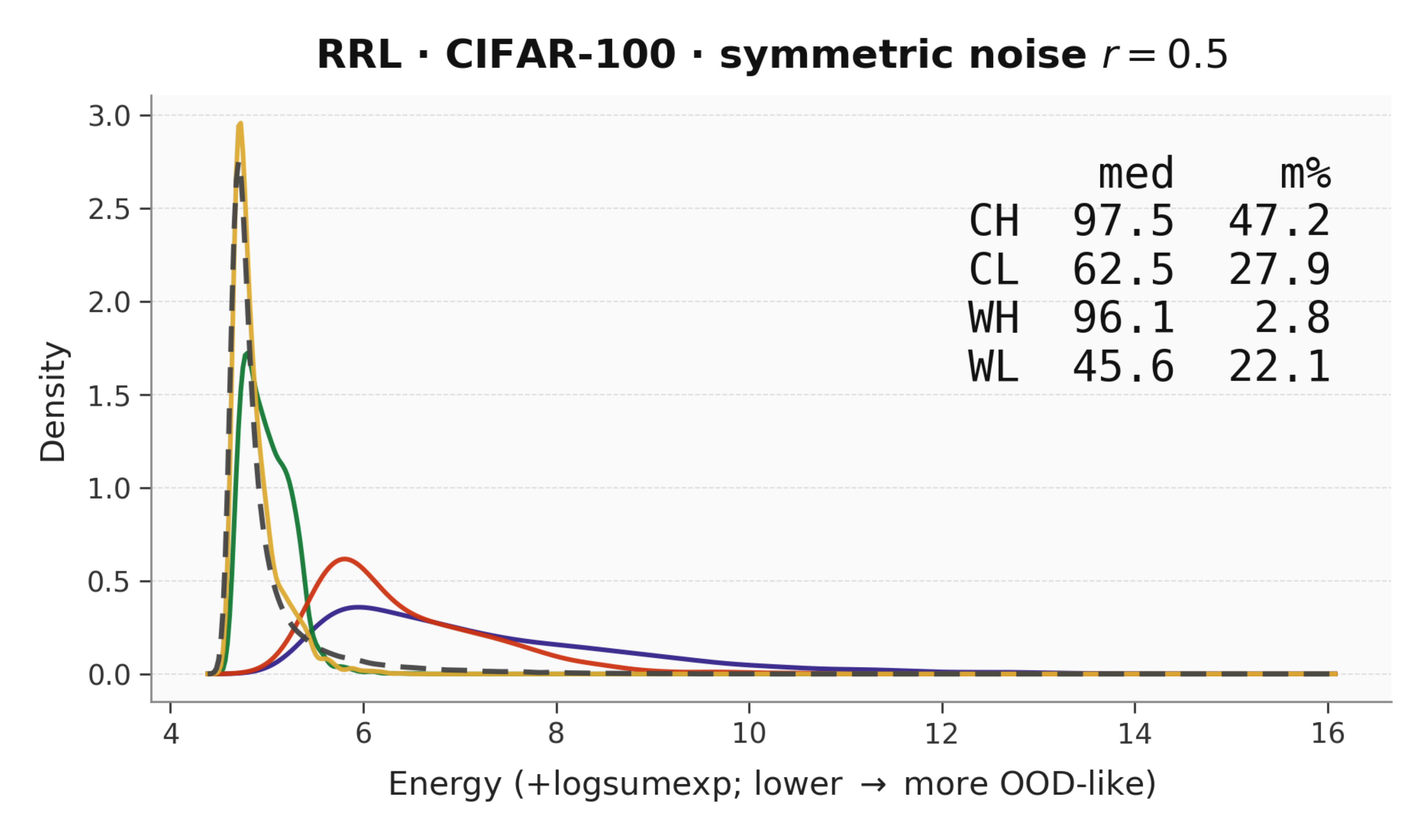} \\
\end{tabular}%
}
\captionof{figure}{\textbf{RRL Energy densities} (kernel-smoothed) on ID-correct, ID-wrong, and pooled far-OOD test points across the seven synthetic-noise settings of Table~\ref{tab:main-benchmark}.
The top-left tile is the shared legend; the remaining panels show RRL under CIFAR-10 symmetric/asymmetric noise and CIFAR-100 symmetric noise.
The low-confidence ID-wrong density repeatedly moves toward the OOD Energy band relative to the ID-correct anchor, illustrating the distributional layer of uncertainty collapse.}
\label{fig:energy-dist-grid}
\endgroup

\paragraph{Interpretation: loss of class-specific structure near boundaries.}
We interpret these signatures as a loss of class-specific structure under noisy labels.
For samples in the low-confidence stratum, the observed geometry is consistent with gradient conflict that may cause the model to discount class-specific features rather than preserve a stable semantic identity.
Consequently, these points lose class-specific structure and slide toward class-transition regions---the geometric locations where OOD features naturally concentrate in angular terms.
In effect, the model places these ID-wrong samples in boundary regions, making them difficult to distinguish from true anomalies in both feature and score space.
Section~\ref{sec:intervention} introduces VMR as a boundary-targeted repair that does \emph{not} relabel these instances; instead it synthesizes virtual outliers on trusted ID batches and enforces energy separation, aiming to widen the margin between this collapsed region and true OOD at fixed post-hoc scores.


\section{Plug-in Repair: Virtual Margin Regularization (VMR)}
\label{sec:intervention}

\paragraph{VMR as a repair probe.}
Motivated by Section~\ref{sec:pathology}, we use \textbf{Virtual Margin Regularization (VMR)} as a diagnostic repair probe rather than as a new noisy-label learning framework.
The intervention asks a narrow question: under the same host learner and the same post-hoc Energy evaluation contract, can widening trusted-ID boundary margins partially recover far-OOD separation?
VMR therefore keeps $\mathcal{L}_{\mathrm{host}}$ unchanged and does not introduce a standalone detector.

\paragraph{What the plug-in changes---and what it does not.}
VMR consumes only the trusted labeled mini-batches already produced by the host LNL algorithm.
It does not mine ID-wrong samples (which are unavailable without clean labels), and it does not filter batches by MSP/disagreement heuristics in the default recipe.
Following VOS~\cite{du2022vos}, VMR synthesizes feature-space virtual outliers near low-likelihood or inter-class boundary regions and applies a lightweight separation term against trusted ID features.
Training uses an additive objective $\mathcal{L}=\mathcal{L}_{\mathrm{host}}+\lambda_{\mathrm{vos}}\mathcal{L}_{\mathrm{VOS}}$, while test-time scoring remains the same post-hoc Energy score in Eq.~\eqref{eq:energy-eval}~\cite{liu2020energy}.
The code-aligned formulation, synthesis rules, and extended sensitivity checks are omitted in this arXiv preprint version.

\paragraph{Structural recovery check.}
Beyond AUROC/FPR95, we verify whether repair changes the diagnosed structure itself.
Concretely, we track collapse diagnostics that measure whether the ID-wrong score distribution moves away from far-OOD relative to the ID-correct anchor under paired baseline/repair checkpoints.
Definitions of $R_{\mathrm{collapse}}$, $\Cscore$, and $M_{\mathrm{collapse}}$, as well as supplementary correlations, are omitted in this arXiv preprint version.

\begin{figure*}[!htbp]
\centering
\IfFileExists{figures/t-sne.pdf}{%
  \includegraphics[width=0.76\linewidth,height=0.30\textheight,keepaspectratio]{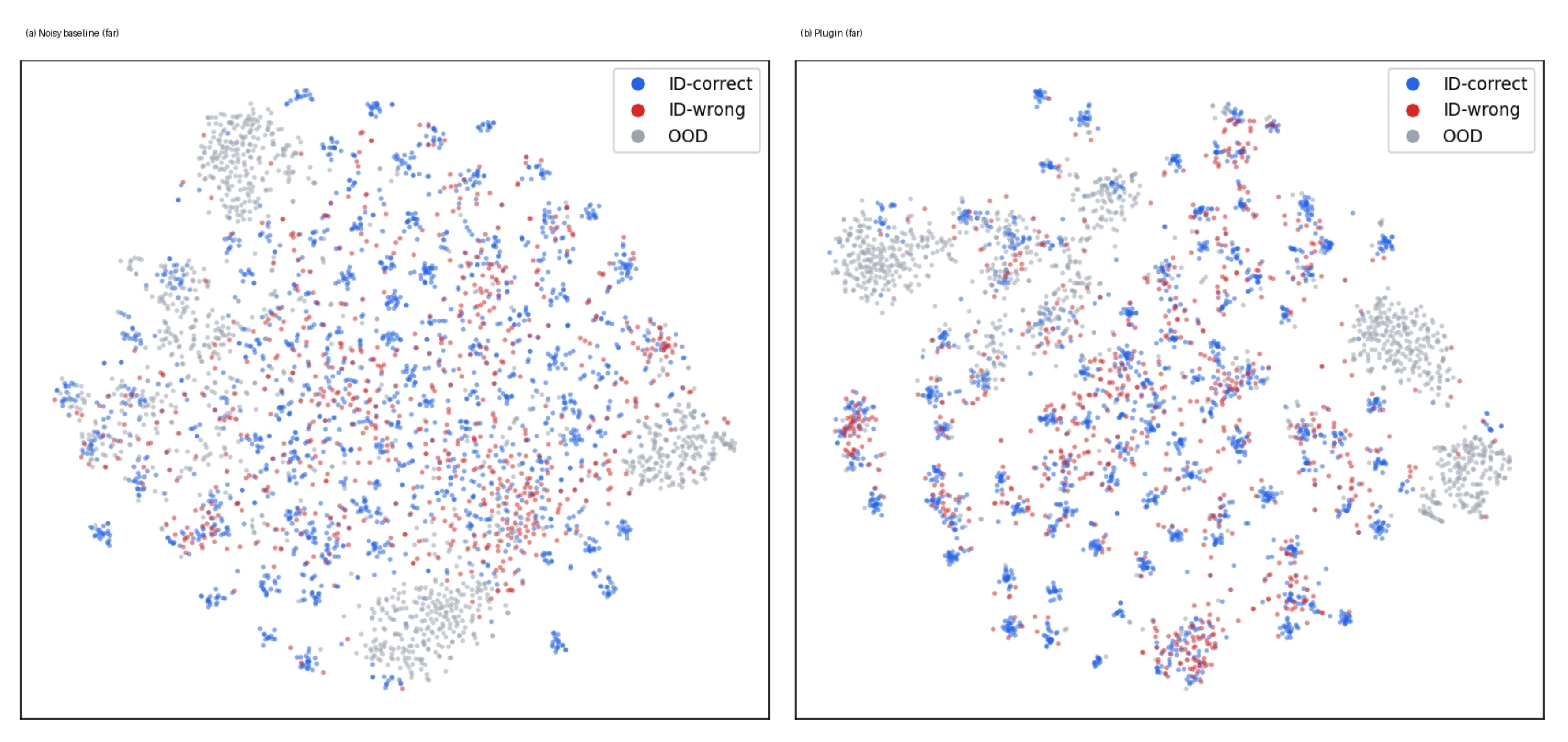}%
}{%
  \fbox{\parbox{0.92\linewidth}{\centering\small Missing figure file:\\
  \texttt{figures/t-sne.pdf}\\[2pt]
  (Feature-probe panel placeholder.)}}%
}
\caption{\textbf{Far-OOD feature probe.}
Two-dimensional projection of penultimate features for UNICON on C100 sym\,$0.5$: ID-wrong (red) bridges ID-correct (blue) and far-OOD (gray), degrading coarse separation in the representation used by post-hoc Energy scoring.
VMR targets this type of bridge; Section~\ref{sec:intervention} reports the resulting metric changes.}
\label{fig:tsne-far-main}
\end{figure*}

\paragraph{Main benchmark effect.}
Figure~\ref{fig:tsne-far-main} visualizes the targeted failure mode: ID-wrong features bridge ID-correct and far-OOD, weakening coarse separation.
Table~\ref{tab:app-acc} then reports matched ACC \textbf{(a)}, far-OOD \textbf{(b)}, and PSSCL+VMR far-OOD lift \textbf{(c)}.
Across the shown CIFAR-10 synthetic rows, PSSCL+VMR retains or improves ACC while improving far-OOD AUROC and FPR95.
This coupled movement supports the diagnosis-to-repair chain: under an unchanged deployment score, boundary-targeted regularization can partially repair the ID-wrong/OOD interface without an observed closed-set trade-off in these rows.

\begin{table*}[!htbp]
\centering
\caption{\textbf{VMR host benchmark and repair summary.}
\textbf{(a)}~closed-set accuracy (\%) and \textbf{(b)}~far-OOD FPR95$\downarrow$ / AUROC$\uparrow$ (\%) on synthetic CIFAR-10 noise; \textbf{(c)}~far-OOD AUROC$\uparrow$ (\%) recovery for PSSCL baseline (BL) vs.\ PSSCL+VMR on completed synthetic and CIFAR-10N rows, with $\Delta$ (far-OOD AUROC) and ID$\Delta$ (closed-set accuracy).
Panels \textbf{(b)} and \textbf{(c)} stack in the right column, flush with \textbf{(a)}.
Columns in \textbf{(a--b)}: C10 sym\,$0.2/0.5/0.8$, C10 asym\,$0.4$.}
\label{tab:app-acc}
\label{tab:app-far-ood}
\label{tab:intervention}
\footnotesize
\setlength{\tabcolsep}{2pt}
\begin{minipage}[t]{0.49\textwidth}
\centering
\textbf{(a) Closed-set accuracy (\%)}\\[0.35em]
\resizebox{\linewidth}{!}{%
\begin{tabular}{@{}lcccc@{}}
\toprule
Method & C10 s0.2 & C10 s0.5 & C10 s0.8 & C10 a0.4 \\
\midrule
Standard CE & 86.8 & 79.4 & 62.9 & 85.0 \\
GCE         & 86.6 & 81.9 & 54.6 & 76.0 \\
GCE+CL      & 90.0 & 89.3 & 73.9 & 78.1 \\
ELR+        & 95.8 & 94.8 & 93.3 & 93.0 \\
MOIT+       & 94.1 & 91.8 & 81.1 & 93.3 \\
Sel-CL+     & 95.5 & 93.9 & 89.2 & 93.4 \\
MixUp       & 95.6 & 87.1 & 71.6 & 77.7 \\
SOP+        & 96.3 & 95.5 & 94.0 & 93.8 \\
BR          & 95.0 & 95.1 & 94.5 & \textbf{94.9} \\
DMLP        & 94.7 & 94.2 & 93.2 & 93.9 \\
\midrule
DivideMix   & 96.1 & 94.6 & 93.2 & 93.4 \\
LongReMix   & 96.3 & 95.1 & 93.8 & 94.7 \\
L2B         & 92.1 & 88.4 & 71.4 & 91.9 \\
PSSCL       & 96.4 & 95.6 & 93.7 & 93.9 \\
UNICON      & 95.1 & 93.7 & 92.1 & 91.7 \\
RRL         & 95.9 & 93.9 & 83.4 & 93.7 \\
\midrule
PSSCL+VMR   & \textbf{97.2} & \textbf{96.3} & \textbf{95.9} & 94.4 \\
\bottomrule
\end{tabular}}
\end{minipage}\hfill
\begin{minipage}[t]{0.49\textwidth}
\centering
\textbf{(b) Far-OOD FPR95$\downarrow$ / AUROC$\uparrow$ (\%)}\\[0.35em]
\resizebox{\linewidth}{!}{%
\begin{tabular}{@{}lcccc@{}}
\toprule
Method & C10 s0.2 & C10 s0.5 & C10 s0.8 & C10 a0.4 \\
\midrule
DivideMix & 64.1 / 69.7 & 51.2 / 77.2 & 67.0 / 77.9 & 93.1 / 49.2 \\
LongReMix & 64.9 / 70.8 & 58.6 / 67.6 & 62.5 / 62.6 & 68.6 / 70.5 \\
L2B       & 61.3 / 74.7 & 64.9 / 75.5 & 86.6 / 69.9 & 66.4 / 78.9 \\
PSSCL     & \underline{24.0} / \underline{93.4} & 26.4 / 92.3 & \underline{31.1} / \underline{87.8} & \underline{33.8} / \underline{87.5} \\
UNICON    & 38.4 / 84.7 & \underline{19.9} / \underline{92.9} & 35.3 / 83.9 & \underline{33.9} / 82.5 \\
RRL       & 84.9 / 66.3 & 98.8 / 51.1 & 99.1 / 30.8 & 98.0 / 41.7 \\
\midrule
PSSCL+VMR & \textbf{18.3} / \textbf{95.8} & \textbf{19.6} / \textbf{95.9} & \textbf{26.3} / \textbf{91.0} & \textbf{22.3} / \textbf{93.7} \\
\bottomrule
\end{tabular}}

\vspace{0.18em}
\noindent
\textbf{(c) PSSCL far-OOD AUROC$\uparrow$ (\%), BL vs.\ VMR}\\[0.12em]
\begin{center}
\scriptsize
\renewcommand{\arraystretch}{0.90}
\resizebox{0.82\linewidth}{!}{%
\begin{tabular}{@{}lcccc@{}}
\toprule
Setting & BL & VMR & $\Delta$ & ID$\Delta$ \\
\midrule
sym\,0.2  & 93.4 & \textbf{95.8} & \textcolor{teal}{+2.4} & \textcolor{teal}{+0.8} \\
sym\,0.5  & 92.3 & \textbf{95.9} & \textcolor{teal}{+3.6} & \textcolor{teal}{+0.7} \\
sym\,0.8  & 87.8 & \textbf{91.0} & \textcolor{teal}{+3.2} & \textcolor{teal}{+2.2} \\
asym\,0.4 & 87.5 & \textbf{93.7} & \textcolor{teal}{+6.2} & \textcolor{teal}{+0.5} \\
\midrule
C10N Aggre & 82.6 & \textbf{96.2} & \textcolor{teal}{+13.6} & \textcolor{teal}{+1.6} \\
C10N Rand1 & 82.6 & \textbf{96.2} & \textcolor{teal}{+13.6} & \textcolor{teal}{+1.6} \\
C10N Worst & 84.0 & \textbf{93.7} & \textcolor{teal}{+9.7} & \textcolor{teal}{+2.5} \\
\bottomrule
\end{tabular}}%
\end{center}
\end{minipage}
\end{table*}

\paragraph{Compact recovery and ablation.}
Panel~\textbf{(c)} of Table~\ref{tab:app-acc} (equivalently Table~\ref{tab:intervention}) summarizes compact far-OOD AUROC gains under VMR.
VMR produces consistent far-OOD gains on completed CIFAR-10 synthetic and CIFAR-10N rows with modest closed-set movement.
Additional ablations are omitted in this arXiv preprint version.
This pattern suggests that repair depends on the host learner's trusted-ID geometry rather than the BCE template alone.
Together, these results connect the benchmark, diagnosis, and intervention: ACC--OOD evaluation exposes the failure, the pathology isolates low-confidence ID-wrong overlap as the diagnostic bottleneck, and VMR shows that virtual-boundary energy separation on trusted batches can partially mitigate that bottleneck without supervised access to misclassified IDs.
The repair remains partial, consistent with far-OOD improving faster than near-OOD under local semantic entanglement.

\section{Discussion and Conclusion}
\label{sec:discussion}

\paragraph{Summary and benchmark suite.}
Noisy-label learners can achieve high closed-set accuracy yet leave a low-confidence misclassified stratum whose post-hoc scores overlap OOD.
This overlap is subpopulation-specific (Table~\ref{tab:five-group}) and is much milder under clean labels than under symmetric noise, supporting a training-dynamics view rather than a purely difficult-image view.
Accuracy-only leaderboards therefore miss a deployment-relevant failure mode.
We recommend that LNL evaluation co-report closed-set generalization and open-world reliability, and we provide a unified benchmark suite that standardizes LNL training, near-/far-OOD routing, post-hoc scoring, ACC--OOD diagnostics, and five-group taxonomy reports.
VMR provides a repair-side check: once the protocol exposes uncertainty collapse, targeted boundary regularization can partially recover far-OOD reliability in the tested rows without sacrificing closed-set accuracy.
Future LNL objectives should therefore be judged by whether they improve the ID-wrong/OOD interface rather than ACC alone.

\paragraph{Limitations and broader impact.}
Our mechanistic evidence is concentrated on CIFAR-family noise with PreAct-ResNet-18 and selected modern baselines, a controlled setting chosen to isolate the optimization mechanisms cleanly; CIFAR-N, Food-101N, Animal-10N, and the RN50 ablation extend the audit but do not exhaust the design space.
Most benchmark entries are single-run measurements, so the reported patterns should be read as broad cross-method trends rather than variance estimates.
Extended backbone, repair-strength, synthesis, labeled-subset, multi-score, and detector-control probes are omitted in this arXiv preprint version; a full audit of distance- and gradient-based OOD paradigms remains future work.
As LNL systems move into domains with long-tail unannotated data, including medical imaging and autonomous driving, understanding and mitigating uncertainty collapse is a step toward deployment reliability.

\bibliographystyle{unsrtnat}
\bibliography{references}

@inproceedings{lee2018cleannet,
  title={CleanNet: Transfer Learning for Scalable Image Classifier Training With Label Noise},
  author={Lee, Kuang-Huei and He, Xiaodong and Zhang, Lei and Yang, Linjun},
  booktitle={Proceedings of the IEEE Conference on Computer Vision and Pattern Recognition (CVPR)},
  month={June},
  pages={5447--5456},
  year={2018}
}

@inproceedings{song2019selfie,
  title={{SELFIE}: Refurbishing Unclean Samples for Robust Deep Learning},
  author={Song, Hwanjun and Kim, Minseok and Lee, Jae-Gil},
  booktitle={Proceedings of the International Conference on Machine Learning},
  year={2019}
}

@techreport{krizhevsky2009learning,
  title={Learning Multiple Layers of Features from Tiny Images},
  author={Krizhevsky, Alex and Hinton, Geoffrey},
  institution={University of Toronto},
  year={2009}
}

@article{song2022lnlsurvey,
  title={Learning from Noisy Labels with Deep Neural Networks: A Survey},
  author={Song, Hwanjun and Kim, Minseok and Park, Dongmin and Shin, Yooju and Lee, Jae-Gil},
  journal={IEEE Transactions on Neural Networks and Learning Systems},
  volume={34},
  number={11},
  pages={8135--8153},
  year={2023}
}

@inproceedings{noisyelephant,
  title={A Noisy Elephant in the Room: Is Your Out-of-Distribution Detector Robust to Label Noise?},
  author={Humblot-Renaux, Galadrielle and Escalera, Sergio and Moeslund, Thomas B.},
  booktitle={Proceedings of the IEEE/CVF Conference on Computer Vision and Pattern Recognition},
  pages={22626--22636},
  year={2024}
}

@inproceedings{wrongline,
  title={Accuracy on the Wrong Line: On the Pitfalls of Noisy Data for Out-of-Distribution Generalisation},
  author={Sanyal, Amartya and Hu, Yaxi and Yu, Yaodong and Ma, Yian and Wang, Yixin and Sch{\"o}lkopf, Bernhard},
  booktitle={Proceedings of The 28th International Conference on Artificial Intelligence and Statistics},
  pages={2170--2178},
  year={2025}
}

@inproceedings{opensetlnl,
  title={Unlocking the Power of Open Set: A New Perspective for Open-Set Noisy Label Learning},
  author={Wan, Wenhai and Wang, Xinrui and Xie, Ming-Kun and Li, Shao-Yuan and Huang, Sheng-Jun and Chen, Songcan},
  booktitle={Proceedings of the AAAI Conference on Artificial Intelligence},
  volume={38},
  pages={15438--15446},
  year={2024}
}

@inproceedings{cifarn,
  title={Learning with Noisy Labels Revisited: A Study Using Real-World Human Annotations},
  author={Wei, Jiaheng and Zhu, Zhaowei and Cheng, Hao and Liu, Tongliang and Niu, Gang and Liu, Yang},
  booktitle={International Conference on Learning Representations},
  year={2022}
}

@inproceedings{dividemix,
  title={{DivideMix}: Learning with Noisy Labels as Semi-Supervised Learning},
  author={Li, Junnan and Socher, Richard and Hoi, Steven C. H.},
  booktitle={International Conference on Learning Representations},
  year={2020}
}

@article{longremix,
  title={{LongReMix}: Robust Learning with High Confidence Samples in a Noisy Label Environment},
  author={Cordeiro, Filipe R. and Sachdeva, Ragav and Belagiannis, Vasileios and Reid, Ian D. and Carneiro, Gustavo},
  journal={Pattern Recognition},
  volume={133},
  pages={109013},
  year={2023}
}

@inproceedings{l2b,
  title={{L2B}: Learning to Bootstrap Robust Models for Combating Label Noise},
  author={Zhou, Yuyin and Li, Xianhang and Liu, Fengze and Wei, Qingyue and Chen, Xuxi and Yu, Lequan and Xie, Cihang and Lungren, Matthew P. and Xing, Lei},
  booktitle={Proceedings of the IEEE/CVF Conference on Computer Vision and Pattern Recognition (CVPR)},
  month={June},
  pages={23523--23533},
  year={2024}
}

@article{psscl,
  title={{PSSCL}: A Progressive Sample Selection Framework with Contrastive Loss Designed for Noisy Labels},
  author={Zhang, Qian and Zhu, Yi and Cordeiro, Filipe R. and Chen, Qiu},
  journal={Pattern Recognition},
  volume={161},
  pages={111284},
  year={2025}
}

@inproceedings{unicon,
  title={{UNICON}: Combating Label Noise through Uniform Selection and Contrastive Learning},
  author={Karim, Nazmul and Rizve, Mamshad Nayeem and Rahnavard, Nazanin and Mian, Ajmal and Shah, Mubarak},
  booktitle={Proceedings of the IEEE/CVF Conference on Computer Vision and Pattern Recognition},
  pages={9676--9686},
  year={2022}
}

@inproceedings{rrl,
  title={Learning from Noisy Data with Robust Representation Learning},
  author={Li, Junnan and Xiong, Caiming and Hoi, Steven C. H.},
  booktitle={Proceedings of the IEEE/CVF International Conference on Computer Vision},
  pages={9488--9497},
  year={2021}
}

@inproceedings{promix,
  title={{ProMix}: Combating Label Noise via Maximizing Clean Sample Utility},
  author={Xiao, Ruixuan and Dong, Yiwen and Wang, Haobo and Feng, Lei and Wu, Runze and Chen, Gang and Zhao, Junbo},
  booktitle={Proceedings of the Thirty-Second International Joint Conference on Artificial Intelligence, {IJCAI-23}},
  pages={4442--4450},
  year={2023},
  doi={10.24963/ijcai.2023/494}
}

@inproceedings{du2022vos,
  title={{VOS}: Learning What You Don't Know by Virtual Outlier Synthesis},
  author={Du, Xuefeng and Wang, Zhaoning and Cai, Mu and Li, Yixuan},
  booktitle={International Conference on Learning Representations},
  year={2022}
}

@inproceedings{guo2017calibration,
  title={On Calibration of Modern Neural Networks},
  author={Guo, Chuan and Pleiss, Geoff and Sun, Yu and Weinberger, Kilian Q.},
  booktitle={Proceedings of the 34th International Conference on Machine Learning},
  pages={1321--1330},
  year={2017}
}

@inproceedings{bendale2016openmax,
  title={Towards Open Set Deep Networks},
  author={Bendale, Abhijit and Boult, Terrance E.},
  booktitle={Proceedings of the IEEE Conference on Computer Vision and Pattern Recognition},
  pages={1563--1572},
  year={2016}
}

@inproceedings{geifman2017selective,
  title={Selective Classification for Deep Neural Networks},
  author={Geifman, Yonatan and El-Yaniv, Ran},
  booktitle={Advances in Neural Information Processing Systems},
  volume={30},
  year={2017}
}

@inproceedings{han2018coteaching,
  title={Co-teaching: Robust Training of Deep Neural Networks with Extremely Noisy Labels},
  author={Han, Bo and Yao, Quanming and Yu, Xingrui and Niu, Gang and Xu, Miao and Hu, Weihua and Tsang, Ivor and Sugiyama, Masashi},
  booktitle={Advances in Neural Information Processing Systems},
  volume={31},
  year={2018}
}

@inproceedings{lee2018mahalanobis,
  title={A Simple Unified Framework for Detecting Out-of-Distribution Samples and Adversarial Attacks},
  author={Lee, Kimin and Lee, Kibok and Lee, Honglak and Shin, Jinwoo},
  booktitle={Advances in Neural Information Processing Systems},
  volume={31},
  year={2018}
}

@inproceedings{hendrycks2017baseline,
  title={A Baseline for Detecting Misclassified and Out-of-Distribution Examples in Neural Networks},
  author={Hendrycks, Dan and Gimpel, Kevin},
  booktitle={International Conference on Learning Representations},
  year={2017}
}

@inproceedings{liang2018odin,
  title={Enhancing The Reliability of Out-of-distribution Image Detection in Neural Networks},
  author={Liang, Shiyu and Li, Yixuan and Srikant, R.},
  booktitle={International Conference on Learning Representations},
  year={2018}
}

@inproceedings{liu2020energy,
  title={Energy-based Out-of-distribution Detection},
  author={Liu, Weitang and Wang, Xiaoyun and Owens, John and Li, Yixuan},
  booktitle={Advances in Neural Information Processing Systems},
  volume={33},
  pages={21464--21475},
  year={2020}
}

@inproceedings{sun2021react,
  title={{ReAct}: Out-of-Distribution Detection with Rectified Activations},
  author={Sun, Yiyou and Guo, Chuan and Li, Yixuan},
  booktitle={Advances in Neural Information Processing Systems},
  volume={34},
  pages={144--157},
  year={2021}
}

@inproceedings{sun2022dnn,
  title={Out-of-Distribution Detection with Deep Nearest Neighbors},
  author={Sun, Yiyou and Ming, Yifei and Zhu, Xiaojin and Li, Yixuan},
  booktitle={Proceedings of the 39th International Conference on Machine Learning},
  pages={20827--20840},
  year={2022}
}

@inproceedings{wang2022vim,
  title={{ViM}: Out-of-Distribution with Virtual-Logit Matching},
  author={Wang, Haoqi and Li, Zhizhong and Feng, Litong and Zhang, Wayne},
  booktitle={Proceedings of the IEEE/CVF Conference on Computer Vision and Pattern Recognition},
  pages={4921--4930},
  year={2022}
}

@inproceedings{zhang2018gce,
  title={Generalized Cross Entropy Loss for Training Deep Neural Networks with Noisy Labels},
  author={Zhang, Zhilu and Sabuncu, Mert R.},
  booktitle={Advances in Neural Information Processing Systems},
  volume={31},
  year={2018}
}

\end{document}